\documentclass[preprint,12pt]{elsarticle}

\usepackage{lineno,hyperref}
\usepackage{bm}
\usepackage{amsfonts}
\usepackage{amssymb}
\usepackage{amsmath}
\usepackage{comment}
\usepackage{isomath}

\usepackage{adjustbox}

\usepackage{scalerel,stackengine}
\stackMath
\newcommand\reallywidehat[1]{%
\savestack{\tmpbox}{\stretchto{%
  \scaleto{%
    \scalerel*[\widthof{\ensuremath{#1}}]{\kern-.6pt\bigwedge\kern-.6pt}%
    {\rule[-\textheight/2]{1ex}{\textheight}}
  }{\textheight}%
}{0.5ex}}%
\stackon[1pt]{#1}{\tmpbox}%
}

\usepackage{stmaryrd} 
\usepackage[usenames]{color}
\usepackage{epsfig}
\usepackage{graphicx}
\usepackage{psfrag}

\graphicspath{{Figures/}}
\usepackage{float}
\floatstyle{boxed}
\newfloat{algorithm}{htb}{loa}
\floatname{algorithm}{Algorithm}
\usepackage{algorithm,algpseudocode}
\usepackage{caption}
\usepackage{subcaption}
\usepackage{color}
\usepackage[table,xcdraw]{xcolor}
\usepackage{multirow}
\usepackage{natbib}
\usepackage{amsmath,amssymb}

\modulolinenumbers[5]

\usepackage[left = 2cm, right=2.5cm, top=2cm, bottom =2cm]{geometry}

\begin{document}
\begin{frontmatter}


\title{Neural-Initialized Newton: Accelerating Nonlinear Finite Elements via Operator Learning}



\author{Kianoosh Taghikhani$^{1,2*}$, Yusuke Yamazaki$^{3}$, Jerry Paul Varghese$^1$,\\ Markus Apel$^{1}$, Reza Najian Asl$^{4}$, Shahed Rezaei$^{1*}$}
\address{$^1$ACCESS e.V., Intzestr. 5, D-52072 Aachen, Germany}
\address{$^2$Institute of Applied Mechanics, \\ RWTH Aachen University, Aachen, Germany}
\address{$^3$Graduate School of Science and Technology, \\ Keio University, Hiyoshi3‑14‑1,\\ Kohoku‑ku, Yokohama 223‑8522, Japan}
\address{$^4$ Technical Universtiy of Munich} 
\address{$^*$ corresponding authors: k.taghikhani@access-technology.de, kianoosh.taghikhani@rwth-aachen.de \\s.rezaei@access-technology.de}

\begin{abstract}
We propose a Newton-based scheme, initialized by neural operator predictions, to accelerate the parametric solution of nonlinear problems in computational solid mechanics. First, a physics-informed conditional neural field is trained to approximate the nonlinear parametric solution of the governing equations. This establishes a continuous mapping between the parameter and solution spaces, which can then be evaluated for a given parameter at any spatial resolution.
Second, since the neural approximation may not be exact, it is subsequently refined using a Newton-based correction initialized by the neural output.
To evaluate the effectiveness of this hybrid approach, we compare three solution strategies: (i) the standard Newton–Raphson solver used in NFEM, which is robust and accurate but computationally demanding; (ii) physics-informed neural operators, which provide rapid inference but may lose accuracy outside the training distribution and resolution; and (iii) the neural-initialized Newton (NiN) strategy, which combines the efficiency of neural operators with the robustness of NFEM. The results demonstrate that the proposed hybrid approach reduces computational cost while preserving accuracy, highlighting its potential to accelerate large-scale nonlinear simulations.



\end{abstract} 

\end{frontmatter}

\section{Introduction}

The numerical solution of nonlinear partial differential equations (PDEs) lies at the heart of computational mechanics and many branches of engineering and applied sciences. The nonlinear finite element method (NFEM) has been the method of choice due to its versatility in handling complex geometries, heterogeneous materials, and multiphysics couplings \cite{belytschko2013nonlinear, bathe1996finite}. In particular, NFEM provides a systematic and mathematically rigorous framework for approximating nonlinear PDEs. However, the main limitation of NFEM lies in its computational cost: nonlinear problems typically require iterative schemes such as the (quasi) Newton–Raphson method, which may require a large number of total iterations. As a result, the computational burden can become prohibitive  \cite{WU2020112704, CHEN2024234054, RUAN2023105169}.

Another important limitation of classical numerical solvers is their one-time-use nature. Specifically, whenever an input parameter, mesh topology, or boundary condition is modified, the solution changes accordingly, requiring the entire computation to be repeated from scratch. This dependency makes their direct use in design and optimization strategies computationally expensive and, in some cases, impractical.

A similar limitation can also be observed in certain deep learning (DL) approaches, such as physics-informed neural networks (PINNs) \cite{karniadakis2021physics}. Moreover, while PINNs have shown promise in some problems in solving forward problems, their application becomes challenging, where the resulting solution fields are expected to exhibit high frequency, oscillatory, and discontinuous behavior \cite{REZAEI2022PINN, WANG2022why, krishnapriyan2021characterizing}.

In recent years, neural operators (NOs) have emerged as a promising paradigm for solving PDEs by learning mappings between function spaces \cite{Lu2021, li2020fourier, kovachki2023neural}. Once trained, NOs can provide solutions to new parameter instances with near-instantaneous inference time, making them attractive for uncertainty quantification, design optimization, and digital twins (see \cite{Rezaei2025npj} and reference therein). Their main strength lies in their ability to generalize within the training distribution and bypass classical discretization schemes. In what follows, we shall briefly review some of the well-known architectures.

Grounded in the universal approximation theorem for operators, the Deep Operator Network (DeepONet) was proposed in \cite{Lu2021}. Since its introduction, DeepONet has been employed across diverse applications, such as predicting nonlinear material response \cite{koric2024deep, he2023novel, GOSWAMI2022114587}. To further enhance its predictive capabilities, multiple improvements have been suggested, ranging from architectural modifications to tailored training strategies \cite{wang2022improved, haghighat2024deeponet, LU2022114778, kontolati2024learning}. 

Another widely studied family of neural operators is the Fourier Neural Operator (FNO), which parameterizes kernels in Fourier space \cite{li2020neural, kovachki2023neural}. By exploiting the efficiency of the Fast Fourier Transform (FFT), FNO achieves strong performance, but its original formulation inherently confines it to rectangular domains. See also other extensions such as Geo-FNO \citet{li2023geofno}.
FNOs have since been adapted to a wide range of engineering applications. For example, \citet{Mehran2022, YOU2022115296, wang2024homogenius} applied FNO to predict complex material behavior, primarily in the elastic regime, using data-driven approaches that rely on offline computations for dataset generation. See also \cite{ESHAGHI2025117785, li2023physicsinformed} for the extension to physics-informed NOs. Later, in \citet{HARANDI2025106219}, a physics-informed extension of FNO was introduced for linear and non-linear micromechanical problems, where FFT was utilized to construct loss functions. This approach, which does not require prior data generation, enables efficient training and accurate predictions in linear, non-linear, and three-dimensional scenarios. 

Nevertheless, neural operators still face several important challenges:
\begin{itemize}
    \item (a) \textit{Data requirement:} Despite their promise, these approaches rely heavily on data, often requiring extensive offline simulations to generate training sets. This step is computationally expensive and limits scalability. It is worth mentioning that the physics-informed versions of NOs tend to solve this issue.
    \item (b) \textit{Generalization to unseen samples:} Normally, it is impractical to cover all possible cases during training. Unless the focus is restricted to a specific application domain, predictions for out-of-distribution inputs can degrade substantially. Although this issue is less pronounced in physics-informed versions, they also suffer from extreme generalization.
    \item (c) \textit{Capturing discontinuous solution fields:} This issue is particularly pronounced in highly nonlinear or strongly coupled problems, where solution fields may exhibit sharp gradients or discontinuities. Accurately representing such features requires the model to capture a broad spectrum of frequencies with varying magnitudes, which is inherently challenging due to the spectral bias commonly observed in vanilla versions of DL-based methods, such as neural operators. 
    \item (d) \textit{Spatial derivatives and sensitivities:} While NOs often achieve acceptable accuracy for the primary solution fields, their reliability significantly decreases when evaluating spatial derivatives or sensitivities (i.e., derivatives of outputs with respect to inputs). These quantities frequently exhibit larger errors, which undermines their utility in engineering applications that rely on accurate sensitivity information. Yet, we know these quantities play a major role in design-oriented tasks.
\end{itemize}
   

Taken together, NFEM and NOs offer highly complementary strengths: NFEM is robust and accurate but computationally expensive, whereas NOs provide fast inference, although sometimes at the cost of accuracy and reliability. Here, the one-time training cost of NOs is not emphasized, as recent physics-informed approaches omit the data-generation part. This complementarity motivates a hybrid strategy that leverages the advantages of both methods. For applications requiring high accuracy, rather than replacing FEM with NOs, the two can be integrated to achieve both high accuracy and computational efficiency.

This integration has manifested itself in various approaches, giving rise to a very recent research direction that is currently under active exploration. Here, we briefly review some key contributions. Broadly, the first category leverages FE knowledge directly in the training of DL models, such as neural operators. In contrast, the second maintains a separation between the two, coupling them in a meaningful way only where necessary.

FEM has been fused with physics-informed operator learning, where \citet{Rezaei2024finite} and \citet{najian_iFOL} used FEM to compute the gradients needed for the physical loss function of a neural network (see also \cite{Rezaei2024fol_mech, Rezaei2025npj}). Similarly, \citet{yamazaki2024} addressed the transient heat equation through a similar approach where the DL framework maps the initial temperature field to the subsequent one by minimizing the residual formulated using FEM. See also similar investigations by \citet{XIONG2025117681} and \cite{LEDUC2023103904}.
\citet{li2025finitepinn} introduces an architecture, which incorporates finite geometric encoding and hybrid Euclidean–topological solution spaces, enabling accurate forward and inverse predictions while embedding physical laws. To address issues with spectral bias, \citet{Moseley2023} introduces finite basis PINNs, which decompose the solution into compact, learned basis functions over overlapping subdomains.

As mentioned, one can view deep learning and finite element methods (or even other direct numerical solvers) as two separate blocks that are coupled together for different purposes. In contrast, within each block, the traditional and well-established approaches are preserved.
A very related work to our current contribution is proposed by \citet{Oommen2024}, who introduced a hybrid framework that couples direct numerical solvers with a U-Net-based neural operator to accelerate materials simulations. 
\citet{THEL2024117073} proposed a framework that accelerates crash simulations by replacing large portions of the FEM mesh with neural networks directly embedded in the solver. Similarly, \citet{PANTIDIS2023115766} introduced a framework that embeds PINNs into the finite element stiffness function to efficiently compute element-level state variables and their derivatives in nonlinear computational mechanics problems. Related approaches can be found in \cite{rezaei2023learning}, where DL predicts the evolution of state variables and hands it over to the nonlinear FE solver.

It is also worth noting efforts to couple DL and FE for accelerating multiscale analyses. \citet{Kalina2023} presented a data-driven multiscale framework combining physics-constrained neural networks as macroscopic surrogate models with FE. Other notable studies include \citet{eivazi2025}, which couples neural operators with FE, and \citet{mianroodi2022lossless}, which integrates U-Net with FE to bridge atomistic and continuum scales.

In this work, we aim to explore the integration of FE and DL from a new perspective. We propose a novel NO$\rightarrow$NFEM strategy in which predictions from a pretrained physics-informed or data-driven neural operator are used as the initial guess for the Newton–Raphson solver in nonlinear FEM. In addition, we exploit the zero-shot super-resolution property of the neural operator by training it on coarser discretizations, effectively reducing the dimensionality of the input space. We demonstrate that the Newton iteration requires only a single load step to reach the exact FEM solution.
A similar line of research has been pursued in several recent studies.
\citet{Kopanicakova2025} proposed a DeepONet-based strategy to accelerate convergence of the preconditioned conjugate gradient method for large-scale linear systems (see also \cite{Zhang2024}).
\citet{ZHOU2025113871} discussed a similar framework that accelerates steady-state fluid simulations by using a vector-cloud neural operator to provide physics-consistent warm starts and achieved acceleration in laminar and turbulent flow problems while maintaining high accuracy.
Compared to related works in this direction, we propose using the output of the neural operator as an initialization for the nonlinear Newton solver. Furthermore, our study focuses on the zero-shot super-resolution capability of the NO, primarily applied to nonlinear problems in solid mechanics.


\color{black}
Among the available options, we focus on a recently developed physics-informed operator learning framework, termed iFOL (implicit finite operator learning) \cite{najian_iFOL}. This framework enables integration of the FE residuals into the loss function. The advantages of iFOL over other well-known operator learning architectures are discussed in \cite{najian_iFOL}. In short, its physics-informed, residual-based formulation enables accurate predictions and strong generalization to unseen problems and complex geometries.
See also {FOLAX} \cite{folax2025github} for codes and latest developments. FOLAX merges classical numerical methods with modern scientific machine learning to solve and optimize parametrized PDEs. By formulating a loss from the Method of Weighted Residuals, it embeds discrete numerical residuals directly into backpropagation.
\color{black}

\begin{figure}[H]
  \centering
  \includegraphics[width=0.99\linewidth]{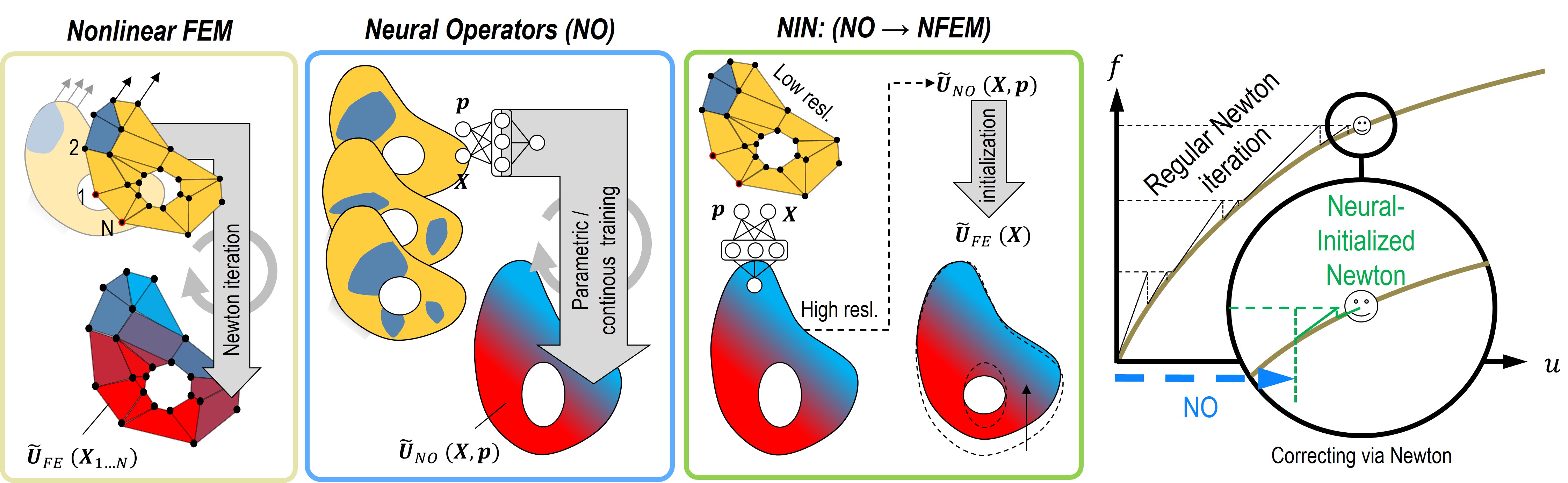}
  \caption{Schematic representation of the three methods investigated in this work.}
\label{fig:idea}
\end{figure}

We compare three approaches, as illustrated in Fig.~\ref{fig:idea}: (A) the standard nonlinear FEM solver, accurate but computationally expensive; (B) neural operators, efficient, yet often less accurate outside the training distribution; and (C) the proposed Neural-Initialized Newton strategy, which combines accuracy and cost-effectiveness. 



\color{black}

\newpage
\section{Formulation and problem setup}
\label{sec:method}
\subsection{Summary of the BVPs}

\color{black}

A broad class of quasistatic problems can be formulated through the minimization of a potential or energy functional $\Pi(\boldsymbol{u},\mathbf{c})$, where $\boldsymbol{u}$ denotes the state field (i.e., displacement or temperature) and $\boldsymbol{c}$ collects problem-specific control parameters (i.e., properties, boundary conditions, and loading). The stationarity condition is expressed in terms of the variational derivative as  
\begin{equation}
    \frac{\delta \Pi(\boldsymbol{u},\boldsymbol{c})}{\delta \boldsymbol{u}} = 0.
\end{equation}
This condition yields the corresponding governing PDE together with the associated boundary conditions. Note that the above condition does not necessarily hold for all problems, nor is it a limiting factor for the present work and the proposed methodologies.
After spatial discretization using the finite element method, the continuous field $\boldsymbol{u}$ is approximated by its nodal values (i.e. $\boldsymbol{U}$), leading to the discrete residual vector at the element level, i.e., $\boldsymbol{r}^e$.  
Upon assembly of all elements (denoted by the operator $\mathcal{A}$), it results in the global nonlinear system of equations  

\begin{equation}
    \boldsymbol{r}(\boldsymbol{U},\boldsymbol{c}) 
    = \mathcal{A}\big[\boldsymbol{r}^e(\boldsymbol{U}^e,\boldsymbol{c})\big] 
    = \boldsymbol{0}.
\end{equation}
Here \( \boldsymbol{U},\boldsymbol{r} \in \mathbb{R}^{M} \) are the vectors of unknowns (discrete solutions at spatial nodes) and nodal residuals, respectively. Again, \( \boldsymbol{c} \) is the control vector that parametrizes the discrete system by governing any or a combination of initial and boundary conditions, material properties, domain geometry, and spatial heterogeneity.

There are various approaches to solve the above equation; here, we focus on the Newton–Raphson algorithm. In this method, the nonlinear residual is linearized, and, given a reasonable initial guess along with a sufficient number of iterations and load steps, the solution can be obtained. The procedure is summarized in Algorithm~\ref{alg:newton_raphson}. It should be noted that, in standard FEM, the solution cannot be directly obtained for arbitrary control parameters \( \boldsymbol{c} \). In other words, the residual vector is minimized only with respect to \( \boldsymbol{u} \) for a given \( \boldsymbol{c} \).

\begin{algorithm}[H]  
\caption{Newton--Raphson Nonlinear Solver with Load Increments} 
\label{alg:newton_raphson}  
\begin{algorithmic}[1]  

\State \textbf{Initialize:} Set total number of load increments \( N_{\text{inc}} \) and \( \bm{u}^{(0)} \gets 0 \).  

\For{load step \( n = 1, \dots, N_{\text{inc}} \)}  

    \State Apply incremental load \( \Delta \bm{f}^{(n)} \).  
    \State Set iteration counter \( k \gets 0 \).  

    \While{convergence not reached}  
    \State \textbf{/* Residual and Tangent stiffness evaluation */}  
        \State $\bm{r}^{(k)} = \bm{f}_{\text{int}}(\bm{u}^{(k)}) - \bm{f}_{\text{ext}}^{(n)}$
        
        \State $\bm{K}^{(k)} = 
        \frac{\partial \bm{r}}{\partial \bm{u}} \bigg|_{\bm{u}^{(k)}}$ 
        \State \textbf{/* Solve linearized system */}   
        \State $\bm{K}^{(k)} \, \Delta \bm{u}^{(k)} = - \bm{r}^{(k)}$ 
        \State \textbf{/* Update solution */}  
        \State \( \bm{u}^{(k+1)} \gets \bm{u}^{(k)} + \Delta \bm{u}^{(k)} \).  
        \State Update iteration counter: \( k \gets k+1 \).  

    \EndWhile  

    \State Accept converged solution \( \bm{u}^{(n)} \).  

\EndFor  

\end{algorithmic}  
\end{algorithm}

\subsection{iFOL: {i}mplicit Finite Operator Learning}
\label{sec:iFOL_gen}
iFOL is based on implicit neural representations (INRs) which are multi-layer perceptron (MLP) networks that are coordinate-based and parameterized by \( L \) layers of weights \( \bm{W}_i \), biases \( \bm{b}_i \), and nonlinear activation functions \( \sigma_i \), with \( \theta = (\bm{W}_i, \bm{b}_i)_{i=1}^{L} \). These networks model spatial fields as an implicit function that maps spatial coordinates to scalar or vector quantities \(
x \in \mathbb{R}^d \mapsto f_{\theta}(x)\). 
We adopt SIREN \citep{sitzmann2020implicit} as the core INR architecture in our framework. This network utilizes sine activation functions, combined with a distinct initialization strategy.
\begin{equation}
\mathbf{SIREN}({x}) = \bm{W}_L (\sigma_{L-1} \circ \sigma_{L-2} \circ \cdots \circ \sigma_0({x}) )+ \bm{b}_L, \quad \text{with} \quad \sigma_i(\boldsymbol{\eta}_i) = \sin\left(\omega_0 (\bm{W}_i \boldsymbol{\eta}_i + \bm{b}_i)\right)
\end{equation}
Here, \( \boldsymbol{\eta}_0 = \boldsymbol{x} \), and \( (\boldsymbol{\eta}_i)_{i \geq 1} \) represent the hidden activations at each layer of the network. The parameter \( \omega_0 \in \mathbb{R}_+^* \) is a hyperparameter that governs the frequency bandwidth of the network. SIREN requires a specialized initialization of weights to ensure that outputs across layers adhere to a standard normal distribution.

In this context, solution to a parameterized partial differential equation using neural fields can be achieved by conditioning the neural field on a set of latent variables \( \boldsymbol{l} \), which can encode the solution field across arbitrary parameterizations and discretization of the underlying PDEs. By varying these latent variables, we can effectively modulate the neural solution field. \( \boldsymbol{l} \) is typically a low-dimensional vector, and is also referred to as a feature code. See also \cite{xie2022neural}.
iFOL utilizes Feature-wise Linear Modulation (FiLM \cite{perez2018film}), which conditions the neural field in an auto-decoding manner. It employs a simple neural network without hidden layers (i.e., a linear transformation) to predict a shift vector from the latent variables \( \boldsymbol{l} \) for each layer of the neural field network. This yields the shift-modulated SIREN:

\begin{equation}
\begin{aligned}
    u_{\theta, \gamma} ({x},\boldsymbol{l}) &=\text{Decode}(\boldsymbol{l}) = \bm{W}_L \left( \sigma_{L-1} \circ \sigma_{L-2} \circ \cdots \circ \sigma_0 ({x}) \right) + \bm{b}_L, \\ 
     \sigma_i(\boldsymbol{\eta}_i,\boldsymbol{\phi}_i) & = \sin \left( \omega_0 (\bm{W}_i \boldsymbol{\eta}_i + \bm{b}_i + \boldsymbol{\phi}_i) \right), \\
     \boldsymbol{\phi}_i(\boldsymbol{l}) & = \bm{V}_i \boldsymbol{l} + \bm{c}_i.
\end{aligned}
\end{equation}

Here, \(u_{\theta, \gamma} \) is the neural solution field, \( \theta = (\bm{W}_i, \bm{b}_i)_{i=1}^{L} \) and \( \gamma = (\bm{V}_i, \bm{c}_i)_{i=1}^{L-1} \) represent the trainable parameters of the SIREN network, referred to as the Synthesizer, and the FiLM hypernetworks, referred to as the Modulator, respectively.
Building on physics-informed neural networks, we introduce a domain-integrated physical loss function whose variation with respect to the solution field yields the residuals of the governing partial differential equations. Among the functionals commonly used in computational mechanics, the total potential energy functional and the weighted residual functional naturally fulfill this property. Since the energy functional is physics-specific and not always explicitly known, we adopt the well-established Method of Weighted Residuals \citep{crandall1956engineering} to formulate the PDE loss function:
\begin{equation}
\label{eq:cont_loss}
\mathcal{L}_{\mathbf{PDE}} = \int_{\Omega} u_{\theta, \gamma} \mathcal{R} \, d\Omega = 0.
\end{equation}
The neural field \(u_{\theta, \gamma}\) serves as the test function, enforcing the vanishing of the residual in a weak (integral) sense. Applying the chain rule and variational calculus, the gradient of the loss function with respect to the predicted solution is given by:
\begin{equation}
\label{eq:cont_loss_variation}
\begin{aligned}
\delta_{u_{\theta, \gamma}}\mathcal{L}_{\mathbf{PDE}} & = \int_{\Omega}  \mathcal{R} \, d\Omega + \int_{\Omega} u_{\theta, \gamma} \,\delta_{u_{\theta, \gamma}}\mathcal{R} \, d\Omega  = \int_{\Omega}  \mathcal{R} \, d\Omega,
\end{aligned}
\end{equation}
where \(\delta_{u_{\theta, \gamma}}\mathcal{R}\) is zero due to the stationarity of the residual functional with respect to the solution field. To compute the loss function efficiently, we discretize the domain and employ the corresponding discrete residuals as follows:  
\begin{equation}
\label{eq:PDE_loss}
    \mathcal{L}_{\mathbf{PDE}}(\bm{u}_{\theta, \gamma}(t,\bm{x},\bm{l}),\bm{c})  = \sum_{e=1}^{n_{el}}({\bm{u}_{\theta, \gamma}^{e}
})^T\bm{r}^{e},
\end{equation}
Here, \(\bm{u}^e_{\theta, \gamma}\) represents the neural solution field vector evaluated at the mesh points, and the superscript \(e\) indicates that the quantity is evaluated at the element level. Here note that, although the PDE loss presented here is based on the residual, one can also directly utilize the energy functional.

The training and inference of CNFs involve the computation of the latent variables \( \bm{l} \), a step commonly referred to as encoding.  iFOL uniquely encodes the PDE and the underlying physics, rather than the spatial fields. In each training step of iFOL, the latent codes for the sample batch \( \mathcal{B} \) are first derived by minimizing the physical loss with respect to the latent codes in just a few steps of gradient descent:
\begin{equation}
\label{eq:encode_PDE_loss}
\begin{aligned}
\boldsymbol{l}^*(\mathbf{c}_i)&= \text{Encode} (\mathbf{PDE}) \\ &= \arg \min_{\boldsymbol{l}} \mathcal{L}_{\mathbf{PDE}} = \sum_{e=1}^{n_{el}}({\bm{u}_{\theta, \gamma}^{e}
})^T\bm{r}^{e}(\bm{u}^{e}_{\theta, \gamma},\bm{c}^{e}_i),\quad \forall i \in \mathcal{B}
\end{aligned}
\end{equation}

The parameters of the Synthesizer and Modulator networks are optimized using the computed latent codes: 
\begin{equation}
\label{eq:back_propagation_step}
\theta^*, \gamma^* = \arg \min_{\theta, \gamma} \mathcal{L}_{\mathbf{PDE}} = \sum_{e=1}^{n_{el}}({\bm{u}_{\theta, \gamma}^{e}
})^T\bm{r}^{e}(\bm{u}^{e}_{\theta, \gamma},\bm{c}^{e}_i),\quad \forall i \in \mathcal{B}
\end{equation}
Basically, we partition the model into context-specific parameters, which dynamically adapt to individual samples, and meta-trained parameters, which are globally optimized to enable knowledge transfer across diverse contexts. This procedure corresponds to second-order meta-learning, as it implicitly involves second-order derivatives of the loss. Algorithm \ref{alg:combined_iFOL} outlines the approach, where \(\alpha\) is the encoding learning rate and \(\lambda\) controls the training updates of the modulator and synthesizer networks.

In this work, we propose a hybrid solver in which a pretrained operator learning model guides the initialization of the nonlinear FEM analysis (see Algorithm \ref{alg:hfe}). 

The goal is to demonstrate that the proposed solver retains the accuracy of NFEM while achieving improved efficiency, as fewer iterations and load increments are typically required. 

Next, we look into some numerical studies to examine this claim.
We begin by summarizing the formulation and problem setup for the nonlinear examples considered. For each case, we briefly outline the governing equations and finite element discretization, followed by details of the corresponding deep learning model. All models, including the finite element components, are implemented in the JAX framework and made openly available for verification and further development at \href{https://github.com/RezaNajian/folax}{FOLAX} \cite{folax2025github}.

\begin{figure}[H]
  \centering
  \includegraphics[width=0.99\linewidth]{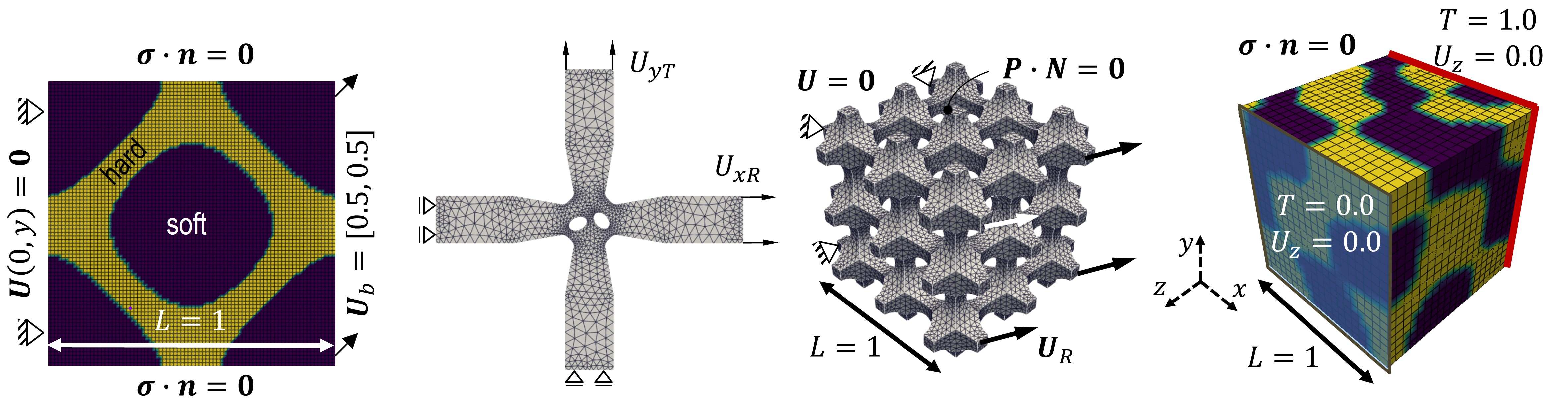}
  \caption{Nonlinear problems investigated in this work. Left: learning deformation patterns for arbitrary material distributions. Middle: learning solution fields on complex geometries under arbitrary loading conditions. Right: learning multiphysics solution fields within a representative volume element.}
\label{fig:examples}
\end{figure}

\color{black}
There are four selected examples in this study, in which the full setup, including boundary conditions, input types to learn the solution, and other key details, is summarized in each section later. The first example concerns the 2D hyperelastic response of a heterogeneous domain, primarily used for detailed analysis. The goal is to predict the solution for arbitrary distributions of soft and hard phases, which form complex composite microstructures. The second and third examples address 3D homogeneous hyperelastic materials with complex geometry subjected to different loading conditions. The fourth example involves a nonlinear thermomechanical problem, where a composite microstructure under a temperature gradient develops mechanical stresses and heat fluxes at saturation. These examples are shown in Fig.~\ref{fig:examples} and are chosen for their practical relevance in materials engineering and are deliberately designed to challenge deep learning models with more practical scenarios.
\color{black}

\begin{algorithm}[H]  
\caption{iFOL Training with PDE Encoding} 
\label{alg:combined_iFOL}  
\begin{algorithmic}[1]  
\State \textbf{Initialize:} Set codes to zero \( \boldsymbol{l}_i \gets 0, \quad \forall i \in \mathcal{B} \);
    \For{each \( i \in \mathcal{B} \) and step \( \in \{1, \dots, K_e\} \)}
        \State \( \boldsymbol{l}_i \gets \boldsymbol{l}_i - \alpha \nabla_{\boldsymbol{l}_i} \mathcal{L}_{\mathbf{PDE}}(\bm{u}_{\theta, \gamma}(t,\boldsymbol{x},\boldsymbol{l}_i),\bm{c}_i) \) \hfill 
    \EndFor \hfill 
    
        \Statex \hrulefill  

\While{no convergence}

    \For{\textbf{each} mini-batch \( \mathcal{M} \subseteq \mathcal{B} \)}

        \State \textbf{/* Encoding step */}
        \State Initialize codes: \( \boldsymbol{l}_i \gets 0, \quad \forall i \in \mathcal{M} \);
        \For{each \( i \in \mathcal{M} \) and step \( \in \{1, \dots, K_e\} \)}
            \State \( \boldsymbol{l}_i \gets \boldsymbol{l}_i - \alpha \nabla_{\boldsymbol{l}_i} 
            \mathcal{L}_{\mathbf{PDE}}(\bm{u}_{\theta, \gamma}(t,\boldsymbol{x},\boldsymbol{l}_i), \bm{c}_i) \);  
        \EndFor  
        \State \textbf{/* Training step */}  
        \State \( \theta,\gamma \gets \theta,\gamma - \lambda \frac{1}{|\mathcal{M}|} 
        \sum_{i \in \mathcal{M}} \nabla_{\theta,\gamma} 
        \mathcal{L}_{\mathbf{PDE}}(\bm{u}_{\theta, \gamma}(t,\boldsymbol{x},\boldsymbol{l}^*_i), \bm{c}_i) \);

    \EndFor  

\EndWhile  

\end{algorithmic}  
\end{algorithm}

\begin{algorithm}[H]
\caption{Neural-Initialized Newton Solver}
\label{alg:hfe}
\begin{algorithmic}[1]
\State \textbf{Input:} Problem parameters $\mathbf{c}$, tolerance $\epsilon$
\State \textbf{/* Step 1: Compute initial guess using trained iFOL */}
\State $\bm{u}_0 \gets \text{Call iFOL and predict solution for } \bm{c} \text{ for desired mesh resolution}$ \Comment{see Algorithm~\ref{alg:combined_iFOL}}
\vspace{0.2cm}
\State \textbf{/* Step 2: Solve nonlinear FEM problem using Newton–Raphson */}
\State $\bm{u} \gets \text{Newton–Raphson}(\bm{r}^e(\bm{u}), \bm{u}_0, \epsilon)$ \Comment{see Algorithm~\ref{alg:newton_raphson}}
\vspace{0.2cm}
\State \textbf{Output:} Converged solution $\bm{u}$
\end{algorithmic}
\end{algorithm}

\noindent
\textbf{Remark 1} In this work, Dirichlet boundary conditions are strictly enforced as hard constraints by modifying the predicted solution field along the Dirichlet boundaries after inference (see \cite{najian_iFOL}). Neumann boundary conditions, on the other hand, are naturally satisfied through the weak formulation of the finite element method. 
\\ \\
\textbf{Remark 2} For simplicity, all dimensions, material properties, and quantities of interest are normalized and reported in dimensionless form. This convention is not repeated throughout the manuscript. Readers may interpret the corresponding dimensional values according to their specific application, as this normalization does not affect the generality of the results.

\color{black}

\subsection{Large deformation 2D elasticity}


\subsubsection{Formulation and problem setup}
\label{sec:formulation_mech2d}

We employ the compressible neo-Hookean model, which is among the most widely used 
constitutive laws for hyperelastic materials. Owing to its simplicity and robustness, 
this model is frequently applied in solid mechanics, biomechanics, and computational 
mechanics studies where large deformations are of interest \cite{ogden1997,holzapfel2000}. 
We define $\boldsymbol{u}$ as the displacement field and $\boldsymbol{F} = \boldsymbol{I} + \nabla \boldsymbol{u}$ 
as the deformation gradient.  
The strain energy density function is given by
\begin{equation}
W_{\text{nonlin}} = \tfrac{\mu(\boldsymbol{X})}{2} \bigl(\bar{I}_1 - 3\bigr) 
+ \tfrac{\kappa(\boldsymbol{X})}{4}\bigl(J_F^2 - 1 - 2 \ln J_F\bigr),
\end{equation}
where $\mu(\boldsymbol{X})$ and $\kappa(\boldsymbol{X})$ denote the shear and bulk moduli, 
$\bar{I}_1 = J_F^{-2/3} I_1$ with $I_1 = \mathrm{tr}(\boldsymbol{C})$ the isochoric 
first invariant of the right Cauchy--Green tensor, and $J_F = \det(\boldsymbol{F})$ 
the Jacobian of the deformation gradient.  
The corresponding first Piola--Kirchhoff stress tensor follows as
\begin{equation}
\boldsymbol{P} = \frac{\partial W_{\text{nonlin}}}{\partial \boldsymbol{F}}
= \mu(\boldsymbol{X}) J_F^{-2/3} 
\left( \boldsymbol{F} - \tfrac{1}{3} I_1 \, \boldsymbol{F}^{-T} \right) 
+ \kappa(\boldsymbol{X}) \left( J_F - \tfrac{1}{J_F} \right) \boldsymbol{F}^{-T}.
\end{equation}

It should be noted that, in the above equation, the deformation gradient tensor in 2D is assumed as follows.

\begin{equation*}
    \bm{F}= \begin{bmatrix}
        F_{11} & F_{12} & 0 \\
        F_{21} & F_{22} & 0 \\
        0 & 0 & 1
    \end{bmatrix}
\end{equation*}

The balance of linear momentum with boundary conditions reads
\begin{align}
\nabla \cdot \boldsymbol{P}(\boldsymbol{X},\boldsymbol{F}) + \boldsymbol{b}_0 &= \boldsymbol{0} 
\quad \text{in } \Omega_0, \\
\boldsymbol{u} &= \bar{\boldsymbol{u}} \quad \text{on } \Gamma_D, \\
\boldsymbol{P}\boldsymbol{N} &= \bar{\boldsymbol{t}} \quad \text{on } \Gamma_N,
\end{align}
where $\boldsymbol{b}_0$ denotes body forces per unit reference volume, 
$\bar{\boldsymbol{u}}$ the prescribed displacement on the Dirichlet boundary $\Gamma_D$, 
and $\bar{\boldsymbol{t}}$ the prescribed traction on the Neumann boundary $\Gamma_N$.  


\subsubsection{DL model to map property to solution field in 2D hyperelasticity}
\label{sec:iFOL}
In this section, we introduce a deep learning model that maps the microstructure topology (i.e., spatial distribution of material properties) to the solution field (i.e., displacement components). Formally, this can be expressed as  
\begin{equation}
    \mathcal{O}: [\mu(\boldsymbol{X}), \kappa(\boldsymbol{X})] \;\mapsto\; \boldsymbol{u}(\boldsymbol{X}).
\end{equation}

As described in~\ref{sec:data}, the heterogeneous material properties are phase-dependent and can be directly determined from the phase value at each integration or nodal point. Therfore, we have 
$\mu(\boldsymbol{X}) = f_\mu\bigl(\phi(\boldsymbol{X})\bigr),~~
\kappa(\boldsymbol{X}) = f_\kappa\bigl(\phi(\boldsymbol{X})\bigr)$, 
where \(\phi(\boldsymbol{X})\) is the phase or parametric field at point \(\boldsymbol{X}\), and \(f_\mu\) and \(f_\kappa\) are mappings from the local phase value to the shear modulus \(\mu\) and bulk modulus \(\kappa\), respectively.
Throughout this work, we employ a linear mapping of the form  
$\mu(\boldsymbol{X}) = a_\mu \, \phi(\boldsymbol{X}) + b_\mu,~~
\kappa(\boldsymbol{X}) = a_\kappa \, \phi(\boldsymbol{X}) + b_\kappa,$
with coefficients chosen such that the input range is normalized between 0 and 1.  
Unless stated otherwise, we set \(a_\mu = a_\kappa = 1\) and \(b_\mu = b_\kappa = 0\).
See Section~\ref{sec:data} for further details.

\begin{figure}[H]
  \centering
  \includegraphics[width=0.99\linewidth]{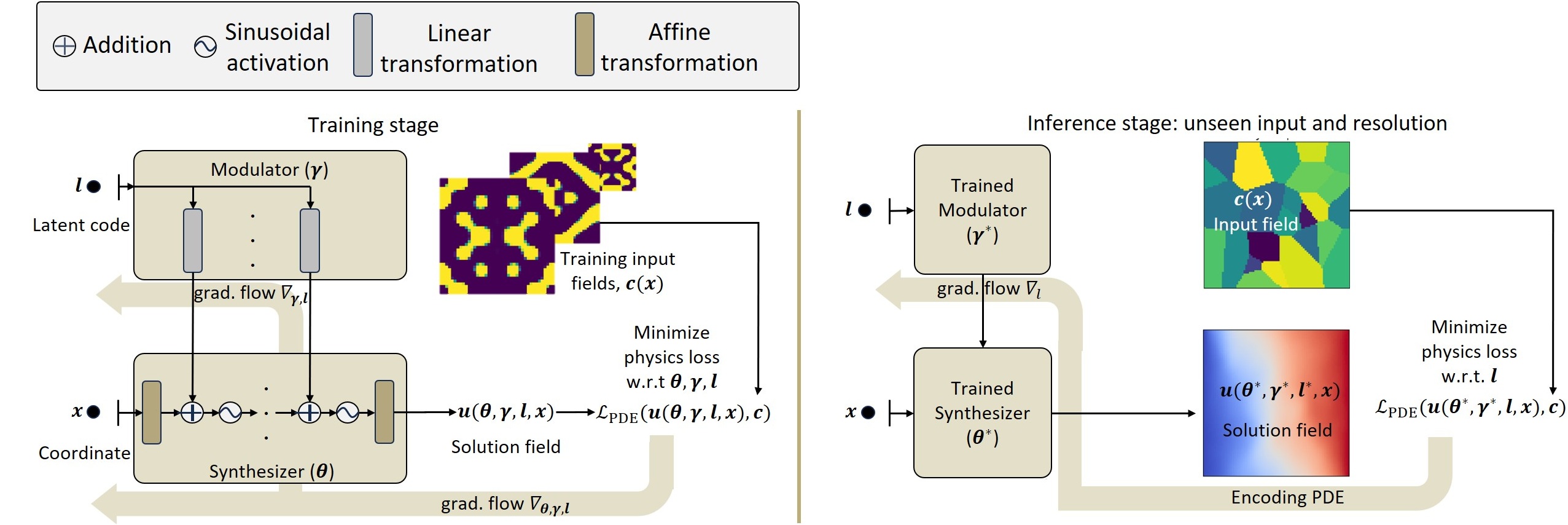}
  \caption{Architecture for physics-informed operator learning to map property fields to solution fields: training on random low-fidelity Fourier samples (left) and inference on realistic high-resolution microstructures (right).}
\label{fig:iFOL_2D_mech}
\end{figure}

The DL model follows Algorithm~\ref{alg:combined_iFOL} together with Eqs.~\ref{eq:encode_PDE_loss} and \ref{eq:back_propagation_step}.
The specific architecture of the proposed neural operator is illustrated in Fig.~\ref{fig:iFOL_2D_mech}.

Following the standard finite element formulation for quadrilateral elements, where the discretized fields are expressed in terms of nodal values and shape functions (e.g., 
$\boldsymbol{u}^e =\boldsymbol{N} \boldsymbol U^e$), the loss function used for training the deep learning model can be written as:
\begin{align}
\mathcal{L} = \sum_{e=1}^{n_{el}} \big(\bm U^e_{\theta, \gamma}\big)^T \bm r^e_u,~~~
\bm r^e_u = 
\sum_{k=1}^{n_{int}} W_k \, \bm B^{nl}(\boldsymbol{\xi}_k,\bm U^e_{\theta, \gamma})^T \hat{\bm S}(\mu(\boldsymbol{\xi}_k,),\kappa(\boldsymbol{\xi}_k) , \bm U^e_{\theta, \gamma}). 
\end{align}

\color{black}
In the above equation, the summations extend over all elements \( n_{el} \) and Gauss integration points \( n_{int} \).  
Here, \( \mu(\boldsymbol{\xi}_k) \) and \( \kappa(\boldsymbol{\xi}_k) \) denote the shear and bulk moduli evaluated at integration point \( \boldsymbol{\xi}_k \). These quantities come from the input set and, in general, are position-dependent as described above. Moreover, \( \boldsymbol{U}^e \) denotes the nodal displacement vector of element \( e \). In the context of physics-informed training, these quantities are treated as unknowns to be inferred through the neural network. 
The weighting factor is given by $W_k = w_k \det(\bm J)$, where $w_k$ are the quadrature weights. The second Piola–Kirchhoff stress is defined as $\boldsymbol{S} = \boldsymbol{F}^{-1} \boldsymbol{P}$, which depends on the information of the material properties at the integration points, and $\hat{\bm S}$ denotes its Voigt notation.
For a brief overview of the finite element formulation of discretized residuals, as well as the definition of $\bm B^{nl}$, the reader is referred to ~\ref{sec:review_FEM}.
\color{black}

\begin{figure}[H]
  \centering
  \includegraphics[width=0.9\linewidth]{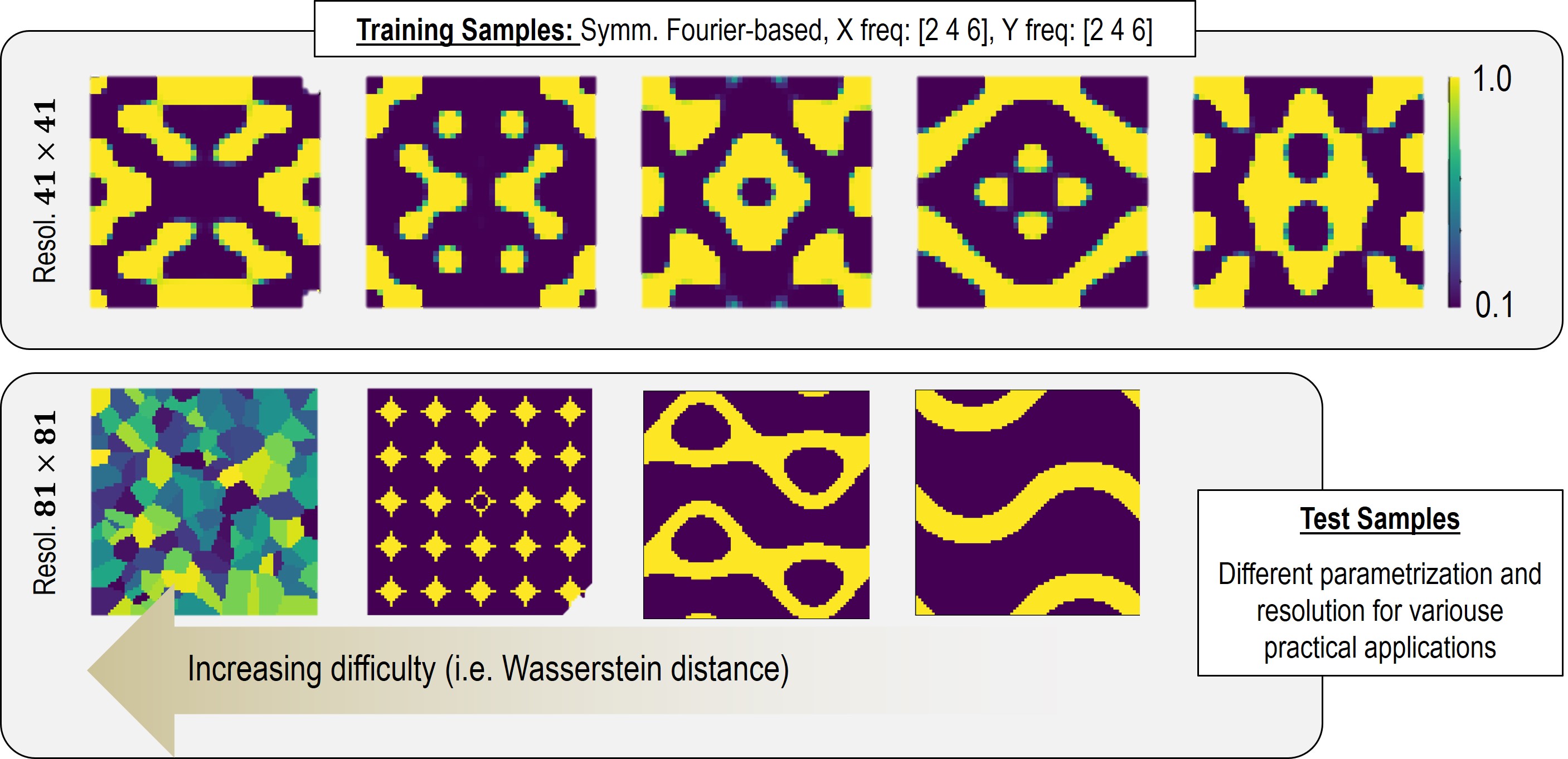}
  \caption{Illustration of a few samples from the training (first row) and test (second row) datasets. Note that during training, the morphologies are limited to low-resolution, Fourier-based, symmetric periodic structures with constant phase contrast, whereas in the test cases we increase the resolution, vary the topologies, and adjust the phase contrast values for more challenging scenarios. }
\label{fig:train_test}
\end{figure}

Training samples (also known as collocation fields) include 8000 binary images, some of which are shown in the first row of Fig.~\ref{fig:train_test}.  
Training samples are generated using the Fourier-based parameterization described in \ref{sec:data}. Training is restricted to fully symmetric and periodic dual-phase samples with fixed phase contrast (that is, \( PC = E_{\max}/E_{\min} =10\)). The resolution is fixed at \(41 \times 41\) for all training samples, which corresponds to \( 1600\) finite element meshes or \(1681\) sensor points. The chosen base resolution is motivated by an initial mesh convergence study with
$PC=10$, as well as by the characteristic frequencies used to generate the topologies, all of which are adequately resolved at this resolution.

\subsection{Large deformation 3D elasticity}
\subsubsection{Formulation and problem setup}
\label{sec:formulation_mech3d}

The formulation remains the same as in the previous section, with the only difference being its extension to 3D.
We selected two examples. The first is inspired by a case of biaxial loading, where the goal is to capture the response of the structure under different applied loads. The second is motivated by loading a metamaterial with complex geometry, where we aim to predict its response directly under given boundary conditions. See the middle part of Fig.~\ref{fig:examples}. 


\subsubsection{DL model to map BCs to solution field in 3D hyperelasticity}
In this study, we investigate a deep learning model that maps the applied loading condition (i.e., Dirichlet BCs on the front surface) to the solution field (i.e., displacement components). Formally, this can be expressed as  
\begin{equation}
    \mathcal{O}: \boldsymbol{U}_b \;\mapsto\; \boldsymbol{u}(\boldsymbol{X}).
\end{equation}

\begin{figure}[H]
  \centering
  \includegraphics[width=0.99\linewidth]{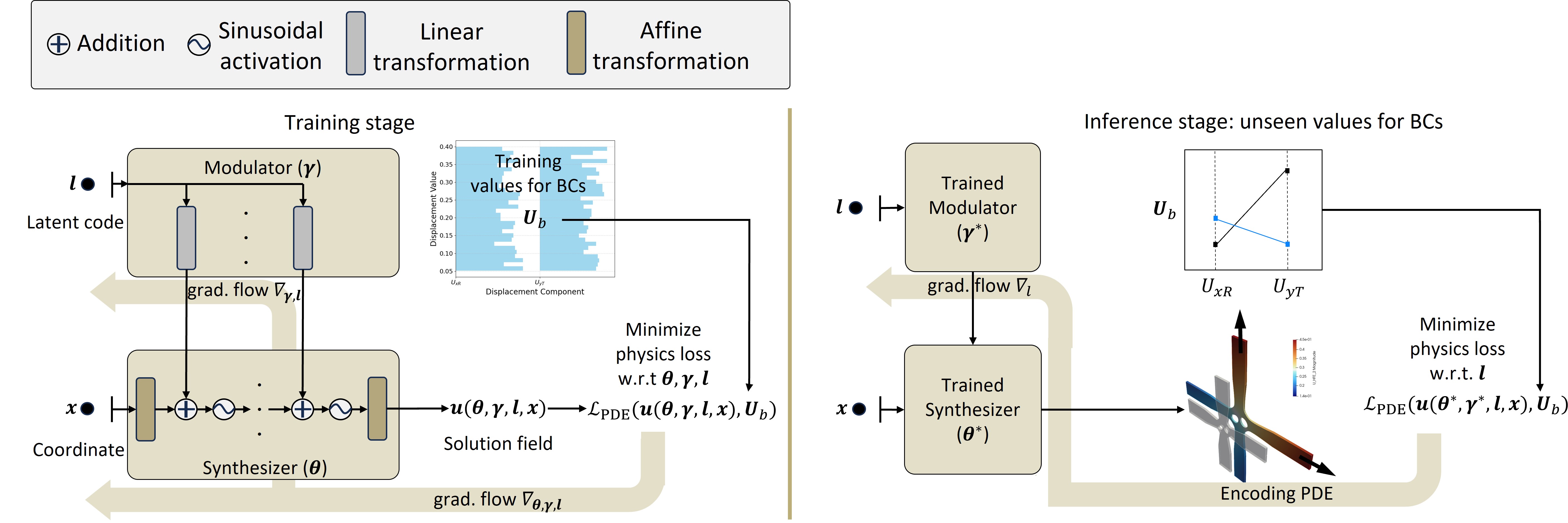}
  \caption{Proposed architecture to map applied Dirichlet boundary conditions to solution fields: training on random loading values (left) and inference on structural responses with complex geometries (right).  }
\label{fig:iFOL_3D_mech}
\end{figure}

\begin{figure}[H]
  \centering
  \includegraphics[width=0.9\linewidth]{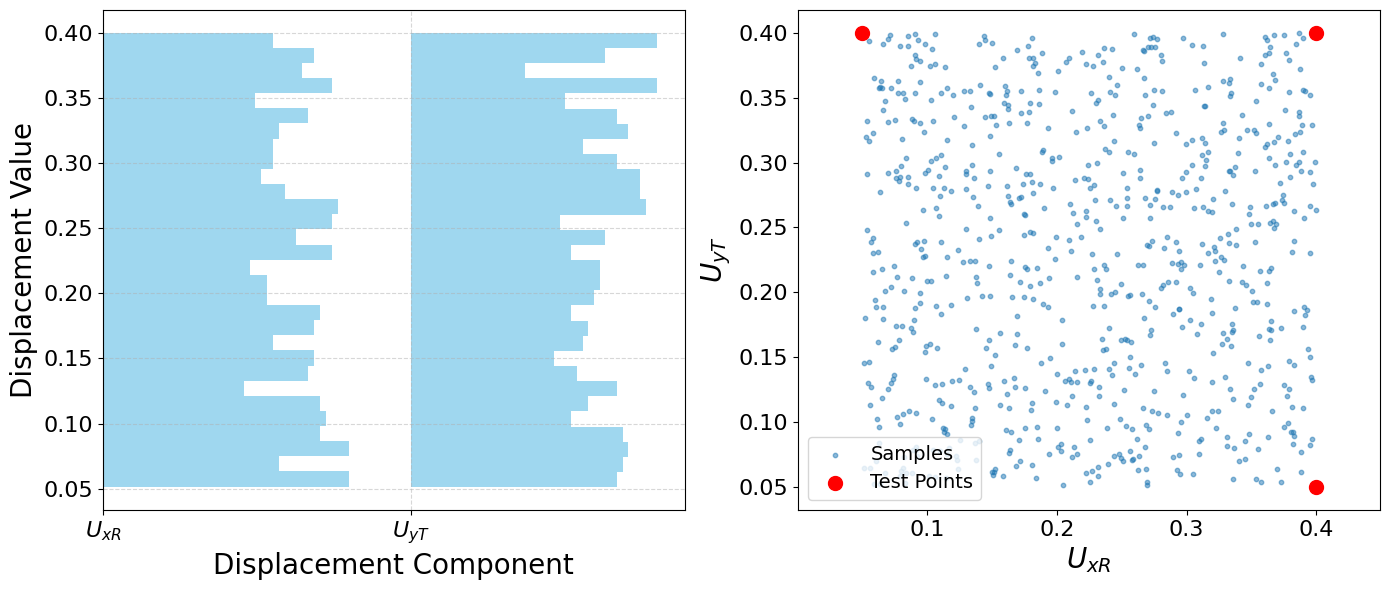}
  \caption{Distribution of Dirichlet boundary values on the top and right edges of the biaxial test. Red dots indicate the selected test cases. }
\label{fig:DBC_dist}
\end{figure}

The architecture of the proposed DL model is shown in Fig.~\ref{fig:iFOL_3D_mech}, while Fig.~\ref{fig:DBC_dist} illustrates the uniform random distribution of the boundary values. From this distribution, three representative extreme test cases are selected for further analysis.

The setup for the second example, involving the learning of nonlinear deformation and stress fields in a complex-shaped metamaterial, follows the same procedure as in the previous figures and is therefore omitted here for brevity. Further details are provided in the results section.

\subsection{Multiphysics problems: Thermo-mechanical system}
\subsubsection{Formulation and problem setup}
\label{sec:formulation_thermomech}

We consider a nonlinear thermomechanical problem involving coupled heat conduction and elasticity with thermal strains. The nonlinearities arise primarily from the temperature dependence of the material properties, while the spatial heterogeneity of the domain follows the setting of the first example. 
The heat transfer is governed by steady-state conduction with position and temperature-dependent conductivity \(k(\boldsymbol{X},T)\):  
\begin{align}
\nabla \cdot  \bm{q}(\boldsymbol{X},T)  &= 0 \quad &&\text{in } \Omega, \\
T &= \bar{T} \quad &&\text{on } \bar{\Gamma}_D, \\
\bm{q} \cdot \bm{n} &= \bar{q} \quad &&\text{on } \bar{\Gamma}_N, \\
\bm{q}(\boldsymbol{X},T) &= - k(\boldsymbol{X},T) \nabla T. &&
\end{align}  
Here, \(T\) denotes the temperature field, \(\bm{q}\) the heat flux vector, \(\bar{T}\) the prescribed temperature on the Dirichlet boundary \(\bar{\Gamma}_D\), and \(\bar{q}\) the prescribed normal heat flux on the Neumann boundary \(\bar{\Gamma}_N\).  
The mechanical equation considering temperature-dependent elastic properties reads:  
\begin{align}
\nabla \cdot \bm{\sigma}(\boldsymbol{X},T) &= \bm{0} \quad &&\text{in } \Omega, \\
\bm{u} &= \bar{\bm{u}} \quad &&\text{on } \bar{\Gamma}_D, \\
\bm{\sigma} \cdot \bm{n} &= \bar{\bm{t}} \quad &&\text{on } \bar{\Gamma}_N,
\end{align}  
where \(\bm{u}\) is the displacement vector, \(\bm{\sigma}\) the Cauchy stress tensor, \(\bar{\bm{u}}\) the prescribed displacement on the Dirichlet boundary, and \(\bar{\bm{t}}\) the prescribed traction on the Neumann boundary.  
For the thermomechanical coupling, the total strain is additively decomposed into elastic and thermal parts, i.e. $\bm{\varepsilon} = \bm{\varepsilon}_e + \bm{\varepsilon}_t$, where we have  
\begin{align}
\bm{\varepsilon}_t &= \alpha \big(T - T_0 \big) \bm{I}.
\end{align}  
Here, \(\bm{\varepsilon}\) is the total strain tensor, \(\bm{\varepsilon}_e\) the elastic strain, \(\bm{\varepsilon}_t\) the thermal strain, \(\alpha\) the thermal expansion coefficient, and \(T_0\) the reference temperature.  
The constitutive relation is expressed by Hooke’s law for heterogeneous linear elasticity:  
\begin{equation}
    \bm{\sigma}(\boldsymbol{X},T) = \mathbb{C}(\boldsymbol{X}) \big( \bm{\varepsilon}(\boldsymbol{X}) - \bm{\varepsilon}_t(\boldsymbol{X},T) \big).
\end{equation}
For isotropic materials, the fourth-order elasticity tensor $\mathbb{C}(\boldsymbol{X})$ can be expressed in terms of the Young’s modulus $E(\mathbf{X})$ and Poisson’s ratio $\nu(\mathbf{X})$ as  
\begin{equation}
    \mathbb{C}_{ijkl}(\boldsymbol{X}) 
    = \frac{E(\boldsymbol{X})}{1+\nu(\boldsymbol{X})} 
    \left( \delta_{ik}\delta_{jl} + \delta_{il}\delta_{jk} \right)
    + \frac{E(\boldsymbol{X}) \, \nu(\boldsymbol{X})}{\big(1+\nu(\boldsymbol{X})\big)\big(1-2\nu(\boldsymbol{X})\big)} \, \delta_{ij}\delta_{kl}.
\end{equation}
where $\delta_{ij}$ denotes the Kronecker delta. This formulation naturally accounts for material heterogeneity by allowing the elastic constants to vary spatially within the domain.

In this study, we introduce the following temperature-dependent thermal conductivity and Young's modulus,
\begin{equation}
    k(T,\boldsymbol{X}) = k_0(\boldsymbol{X})(1+ b~T^{c})
\end{equation}
\begin{equation}
    E(T,\boldsymbol{X}) = E_0(\boldsymbol{X})(1+ b~T^{c})
\end{equation}
where $k_0$ and $E_0$ are the given phase contrast values and $b$ and $c$ control the degree of temperature dependence and non-linearity. In this study, we consider $b=c=2.0$.

\subsubsection{DL model to map property to solution field in 3D nonlinear multiphysics}
\begin{figure}[H]
  \centering  
  \includegraphics[width=0.99\linewidth]{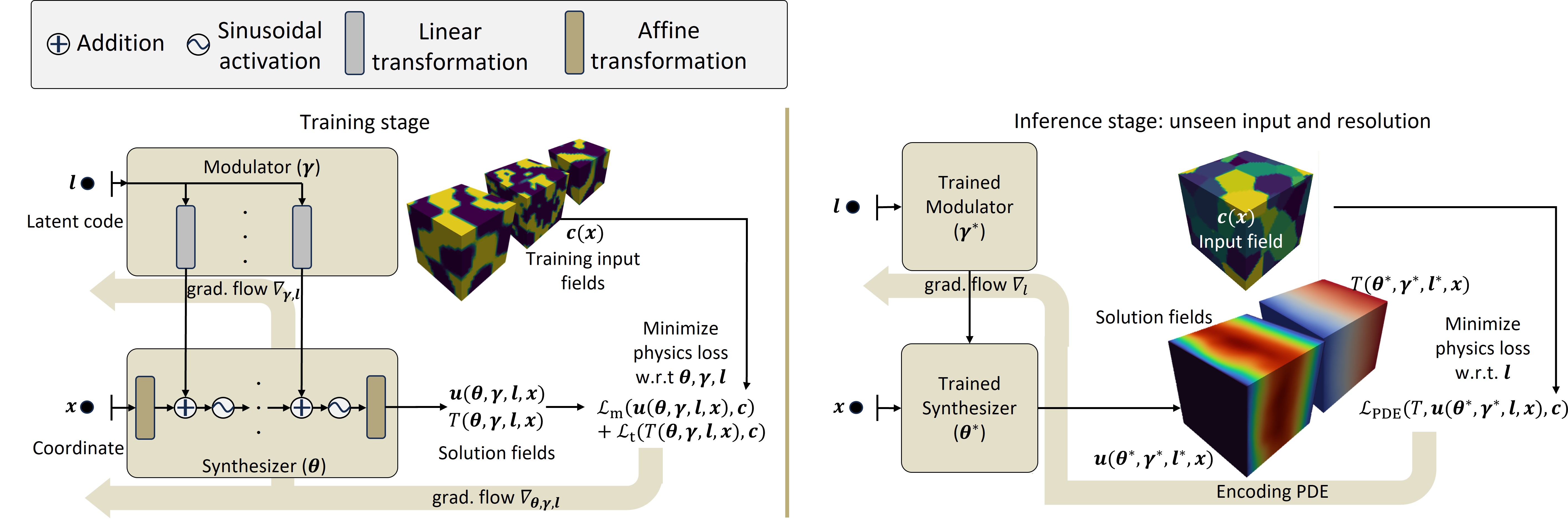}
  \caption{Architecture of a DL model for mapping property fields to solution fields in a nonlinear multiphysics scenario: training on low-fidelity Fourier samples (left) and inference on high-resolution microstructures (right).}
\label{fig:iFOL_3D_thermomech}
\end{figure}

The deep learning model in this case maps the microstructure topology (i.e., spatial distribution of material properties) to the solution field (i.e., displacement and temperature components). Formally, this can be expressed as  
\begin{equation}
    \mathcal{O}: \left[E(\boldsymbol{X}), k(\boldsymbol{X}) \right] \;\mapsto\; \left[\boldsymbol{u}(\boldsymbol{X}), {T}(\boldsymbol{X}) \right].
\end{equation}
Similar to explanations in Section \ref{sec:formulation_mech2d} and \ref{sec:data}, we have $E_0(\boldsymbol{X})=k_0(\boldsymbol{X})=\phi(\boldsymbol{X})$. The specific architecture of the proposed neural operator is illustrated in Fig.~\ref{fig:iFOL_3D_thermomech}.
The coupled thermomechanical problem is enforced by a composite loss function, which accounts for both thermal ($\mathcal{L}_{t}$) and mechanical ($\mathcal{L}_{u}$) contributions:
\begin{equation}
\mathcal{L} = \mathcal{L}_{t} + \mathcal{L}_{u} = \sum_{e=1}^{n_{el}} \big(\bm T^e_{\theta, \gamma}\big)^T \bm r^e_t + \sum_{e=1}^{n_{el}} \big(\bm U^e_{\theta, \gamma}\big)^T \bm r^e_u.
\end{equation}
Here, the thermal and mechanical losses are written for a discretized finite element mesh,
where $\bm r^e_t$ and $\bm r^e_u$ denote the residual vectors of the thermal and mechanical element level, respectively.  
The thermal and mechanical residuals for element $e$ is given by
\begin{align}
\bm r^e_t &= 
\sum_{k=1}^{n_{int}} W_k
\bm B_t(\bm\xi_k)^T 
\bm N_t(\bm\xi_k) k(\bm\xi_k, \bm T^e_{\theta, \gamma}) 
\bm B_t(\bm\xi_k) \bm T^e_{\theta, \gamma} 
- \sum_{k=1}^{n_{int}} W_k \bm N_t(\bm\xi_k)^T \bm f^e, \\
\bm r^e_u &= 
\sum_{k=1}^{n_{int}} W_k \, \bm B_u(\bm\xi_k)^T \bm D(\bm\xi_k, \bm T^e_{\theta, \gamma}) \Big( \bm B_u(\bm\xi_k) \bm U^e_{\theta, \gamma} - \alpha \bm N_t(\bm\xi_k) ( \bm T^e_{\theta, \gamma} - \bm T^e_0) \bm s^e \Big)
- \sum_{k=1}^{n_{int}} W_k \, \bm N_u^T \bm f^e.
\end{align}
In the above equations, $k(\bm\xi_k, \bm T)$ denotes the temperature-dependent thermal conductivity, 
while $\bm N_t$ and $\bm N_u$ are the shape function matrices for the thermal and mechanical fields, respectively. 
We further define the thermal gradient matrix 
$\bm B_t = \left[\tfrac{d \bm N_t}{d \bm X}\right]$ 
and the strain–displacement matrix 
$\bm B_u = \left[\tfrac{d \bm N_u}{d \bm X}\right]$. 
The weighting factor is given by $W_k = w_k \det(\bm J)$, where $w_k$ are the quadrature weights. See \ref{sec:review_FEM} for further details.
In the mechanical residual, $\bm D$ represents the elastic stiffness matrix and $\bm s^e$ denotes the stress coupling vector at the element level. 
Together, these residuals constitute the foundation of the physics-informed loss, enforcing the satisfaction of both thermal and mechanical governing equations in a strongly coupled manner.

The domain of interest is a cube with 
$L_x=L_y=L_z=1~\mu$m (see also Fig.~\ref{fig:examples}). The geometry and boundary conditions are summarized in Table \ref{bc_table1}.
In this example, we modify the internal microstructure topology, i.e., the material distribution and property values, within the domain. This setup is particularly relevant for multiscale material development, especially in additive manufacturing and casting.

\begin{table}[H]
\centering
\caption{List of temperature and displacement boundary conditions. Here, "free" indicates Dirhichlet boundary conditions are not enforced; Instead, natural boundary conditions are applied.}
\begin{tabular}{c|ccccccc}
Location  & $x=0$ & $x=L_x$ & $y=0$ & $y=L_y$ & $z=0$ & $z=L_z$ \\
\hline
$T$  & $0.0 $ & $1.0 $  & free & free & free  & free   \\
$U_x$  & $0.0$ & $0.0 $  & free & free & free  & free   \\
$U_y$ &free & free  & $0.0$ & $0.0$ & free  & free  \\
$U_z$ &free & free & free  & free & $0.0$ & $0.0$ \\
\hline
\end{tabular}
\label{bc_table1}
\end{table}

\begin{figure}[H]
  \centering
  \includegraphics[width=0.99\linewidth]{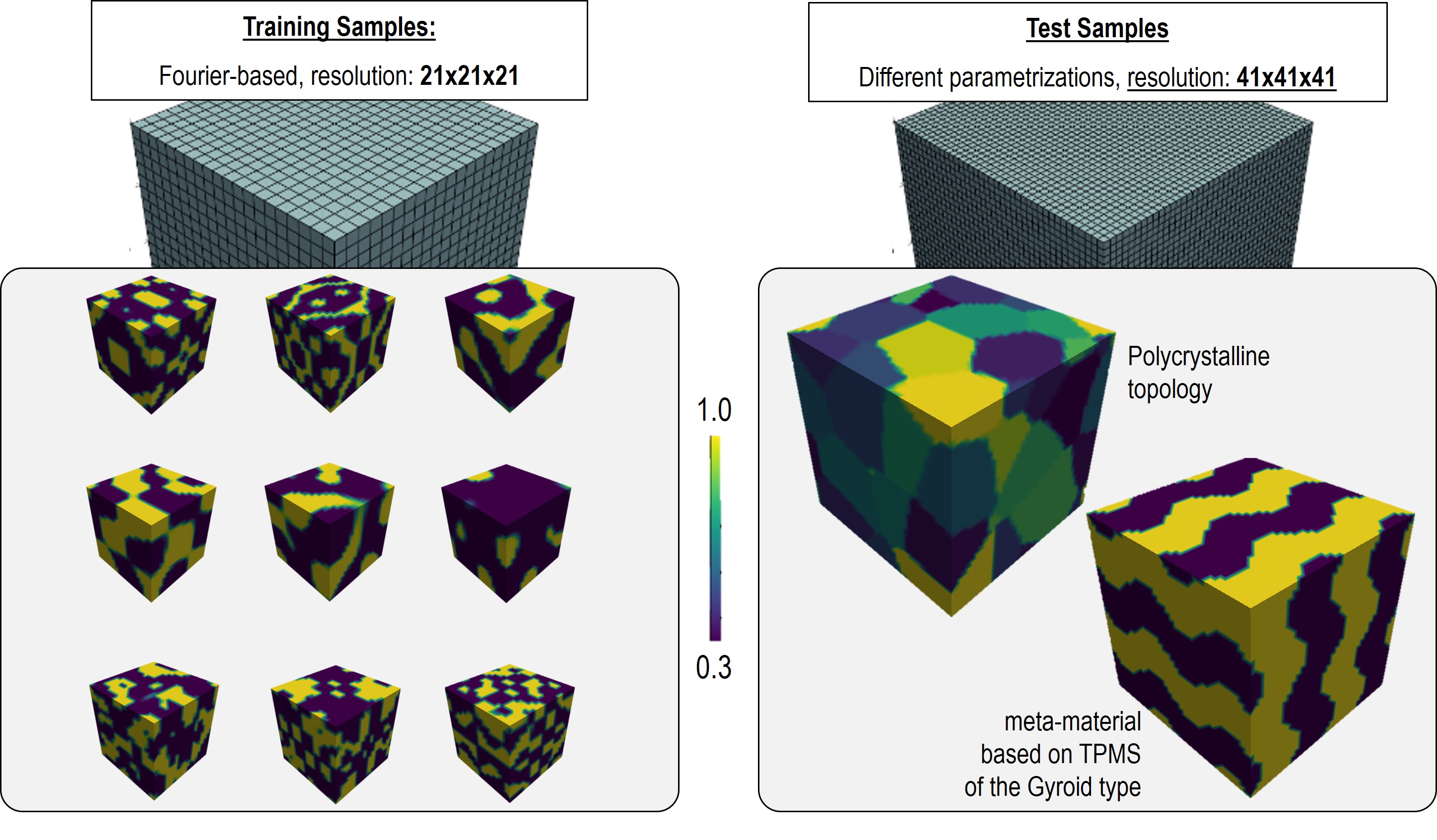}
  \caption{Illustration of a few samples from the training (left) and test (right) datasets. For training, we are limited to low-resolution, Fourier-based structures with constant phase contrast, whereas in the test cases we increase the resolution, vary the topologies, and adjust the phase contrast values.}
\label{fig:train_test_3D_multi}
\end{figure}

The training samples consist of 4000 microstructures that are generated using the Fourier-based parameterization described in \ref{sec:data} and are restricted to dual-phase configurations with a fixed phase contrast, i.e., 
$
PC = \frac{E_{\max}}{E_{\min}} = \frac{1.0}{0.3} \approx 0.33
$. See also left-hand side of Fig.~\ref{fig:train_test_3D_multi}.
Training is performed on a lower-resolution grid of $21 \times 21 \times 21 = 9261$ points, while testing is conducted on a finer grid of $41 \times 41 \times 41 = 68921$ points, corresponding to 64000 finite-element hexahedral meshes. For the fine mesh in a thermo-mechanical coupled setting, this amounts to approximately $0.3$ million degrees of freedom.

The test cases are also intentionally selected based on practical applications.
The TPMS Gyroid, with its smooth, interconnected geometry, combines mechanical strength and efficient heat transfer and it is an ideal test case RVE for exploring thermomechanically coupled behavior in architected materials.
We also consider a polycrystalline test sample generated with the MICRESS software \cite{MICRESS} using a multiphase-field simulation of the austenite-to-ferrite transformation \cite{APEL2009589}. This generic 3D grain structure serves as a benchmark RVE for investigating thermomechanically coupled behavior in multiphase materials.

\color{black}
\section{Results}
\label{sec:results}

In this section, we present the results of the trained operator learning framework (i.e., iFOL) along with the neural initiated Newton (NiN) approach (see Algorithm \ref{alg:hfe}). The reference solution is given by nonlinear FEM (NFEM) simulations (see Algorithm \ref{alg:newton_raphson}). Here the results of NiN essentially coincide with those of NFEM. 
For all reported calculations, the stopping tolerance for the Newton algorithm was set to $1.0e-6$ based on the norm of the residual vector.
Details on training, hardware, and software specifications for NFEM computations, as well as the iFOL framework and hyperparameters used, are provided in \ref{sec:hyperparameters}.

\subsection{Studies on 2D hyper elastic}
\label{sec:2D_elas} 
Our objective is to directly predict the displacement components for an arbitrary microstructure at any resolution. See also the left-hand side of Fig.~\ref{fig:examples} and the formulation discussed in Section \ref{sec:formulation_mech2d}. 
We select 50 unseen test samples, which introduces significant variations in both the input topology (using different parametrization techniques and quantified by an increased Wasserstein distance) and material property values. These include Fourier-based samples with varying initial frequencies, polycrystalline, and dual-phase microstructures derived from TPMS. Moreover, the evaluation is performed at a higher resolution, that is, \(81 \times 81\). 



\begin{figure}[H]
  \centering
\includegraphics[width=0.95\linewidth]{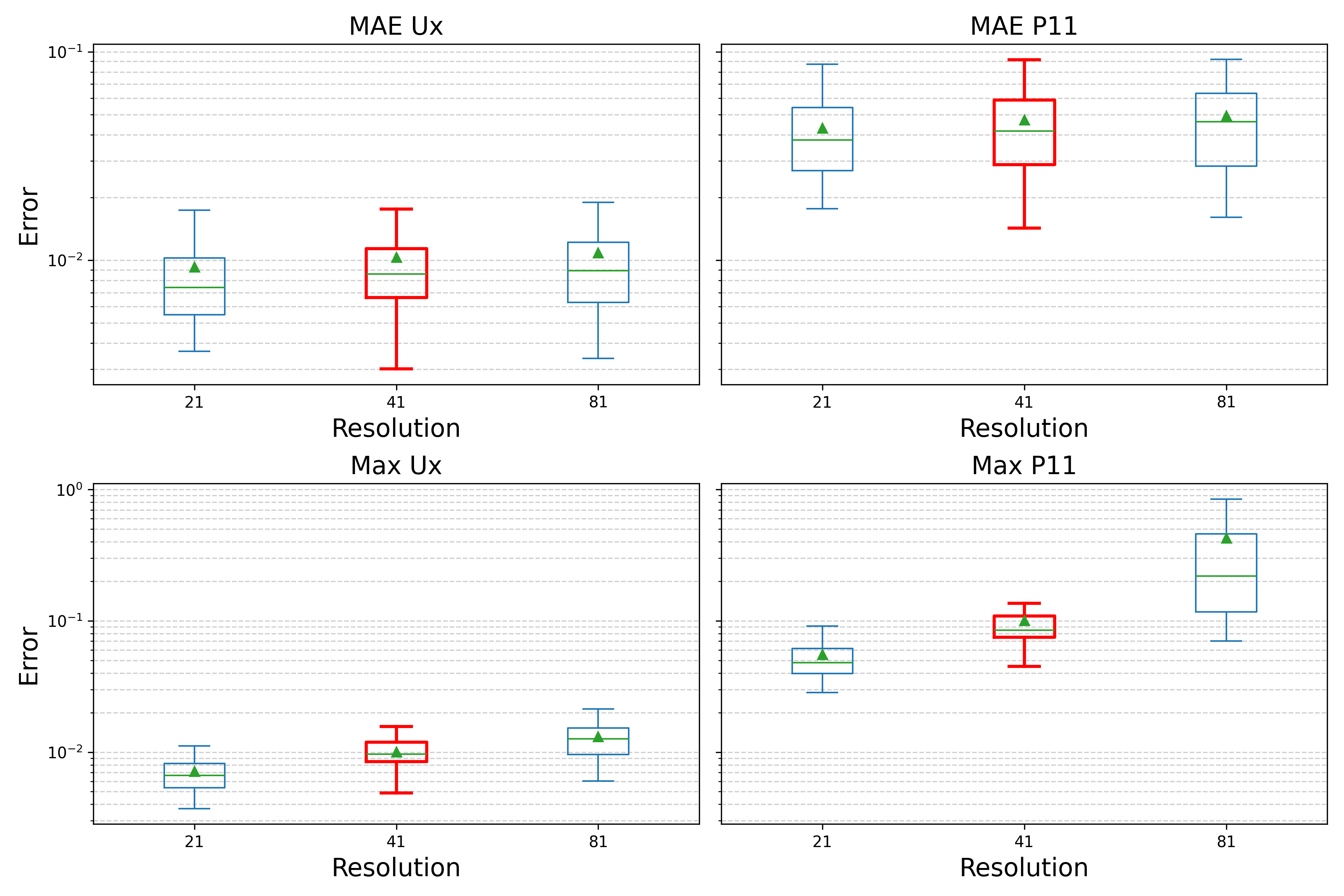}
  \caption{ Error analysis of the iFOL method for test samples at different mesh resolutions. Values marked in red denote the training resolution. }
  
\label{fig:Error_eval_box_plot}
\end{figure}

We start by examining the performance of the trained iFOL, to highlight its unique characteristics and potential shortcomings, which motivate the rest of this work.
In Fig.~\ref{fig:Error_eval_box_plot}, error bars and average values (triangles) are shown for all test samples. It is worth mentioning that although the topology changes significantly, the phase contrast for the 50 test samples considered in this study is kept the same as in training, namely $PC = 10$. 

The iFOL displacement predictions are largely resolution-independent. The maximum point-wise errors $Err_{\max} \;=\; \max_{1 \leq i,j \leq N} \; \bigl|\, \bullet^{\text{FEM}}_{i,j} - \bullet^{\text{iFOL}}_{i,j} \,\bigr|$ are higher, reflecting a more stringent metric, while the mean absolute average error 
$\text{MAE}_{\bullet} = 
\frac{1}{N} \sum_{i=1}^{N}\sum_{j=1}^{N} 
\left| \bullet^{\text{FEM}}_{i,j} - \bullet^{\text{iFOL}}_{i,j} \right|
$ remains acceptable. where $N$ is the total number of nodes (e.g. 21 $\times$ 21 or 41 $\times$ 41 or 81 $\times$ 81).



For stress components, errors increase with higher resolution, highlighting a limitation of current neural operator approaches when evaluating derived quantities. These observations underscore the need for careful error assessment, especially for stress quantities. Stress quantities can also be directly predicted as a remedy \cite{REZAEI2022PINN, HARANDI2025106219, Rezaei2025npj}. 

\begin{figure}[H]
  \centering
\includegraphics[width=0.95\linewidth]{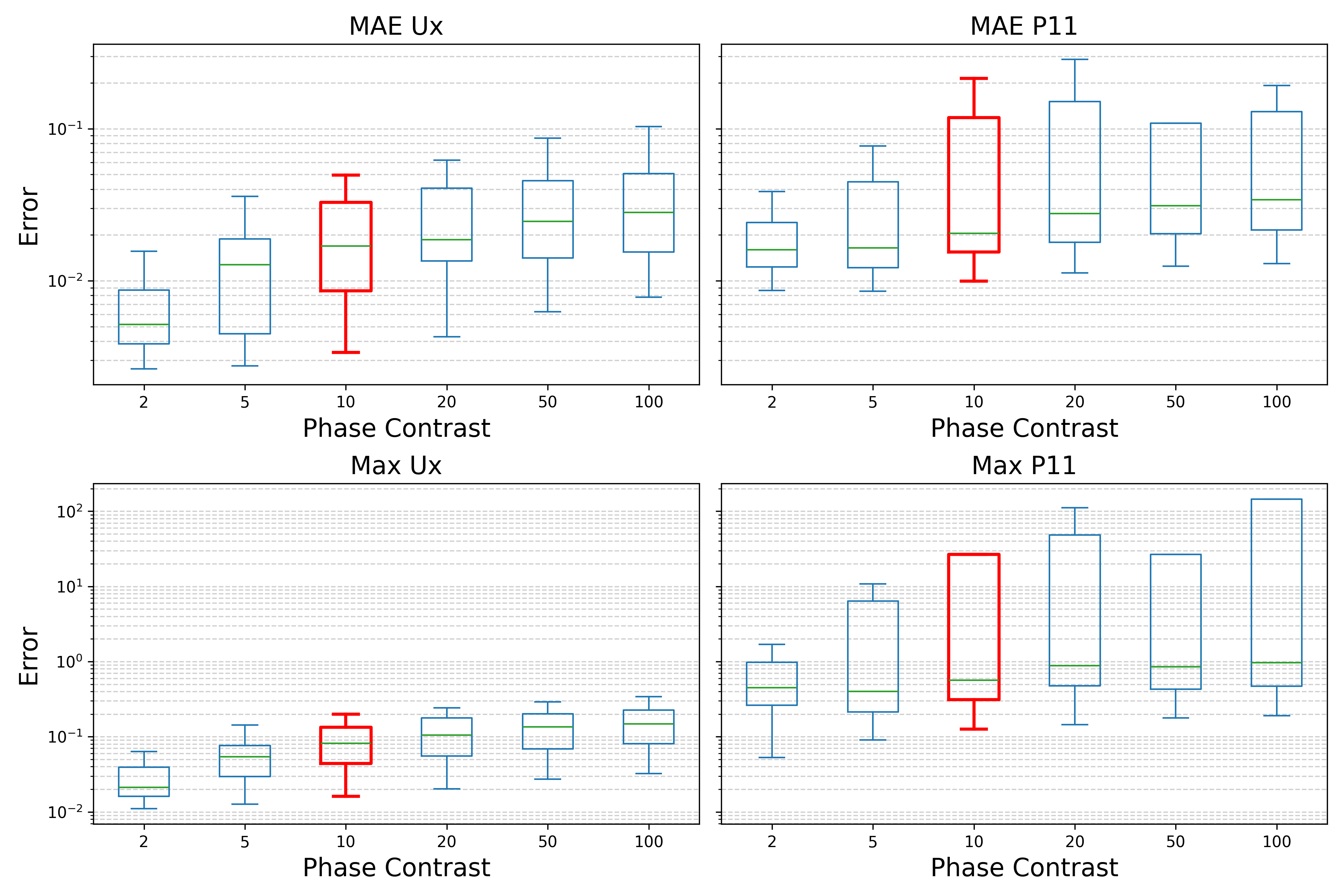}
  \caption{ Error analysis of the iFOL method for test samples with different phase contrast values. Values marked in red denote the training phase contrast. }
\label{fig:PC_Plot}
\end{figure}

Next, we evaluate the same 50 test topologies but with different phase contrast ($PC$) values (see Fig.~\ref{fig:PC_Plot}). Increasing $PC$ leads to sharper jumps and discontinuities, making the problem numerically more challenging and explaining the higher errors. Interestingly, the model performs well, and in some cases even better, for unseen lower $PC$ values. The red lines in Fig.~\ref{fig:PC_Plot} indicate the training $PC$. All results are tested on a higher-resolution $81 \times 81$ grid.

\begin{figure}[H]
  \centering
\includegraphics[width=0.99\linewidth]{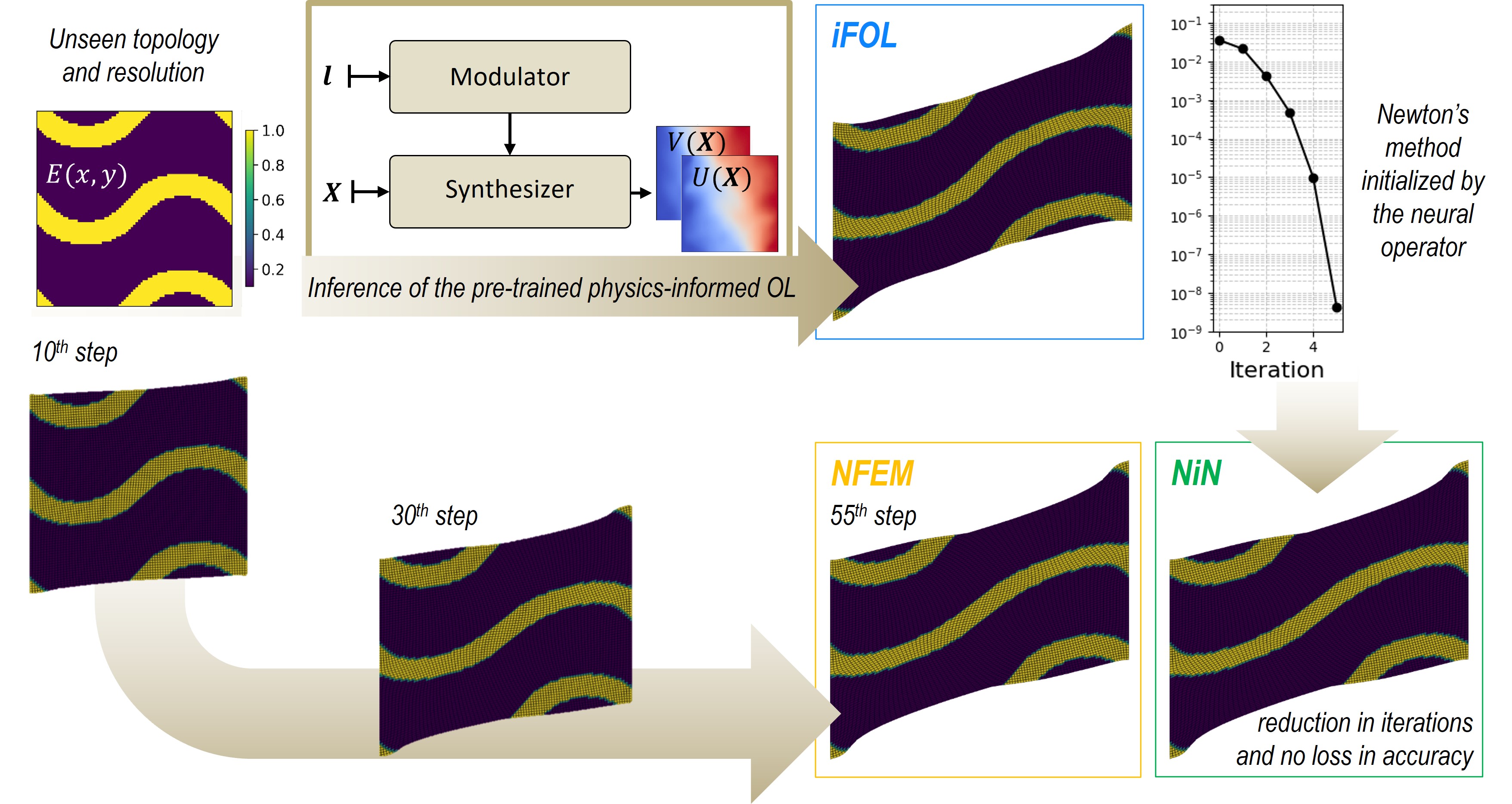}
  \caption{ Performance of the Proposed Neural-Initialized Newton Method Compared to a Surrogate Deep Learning Model and a Standard Nonlinear FE Solver  }
\label{fig:deformed}
\end{figure}

At this point, we discuss how errors can be reduced to the level of direct numerical solver errors on higher-resolution unseen test cases. The first out-of-distribution sample is motivated by TMPS-based structures. 
The proposed methodology is summarized in Fig.~\ref{fig:deformed}. In addition, we report the iterations required for the nonlinear FEM. It is well known that the Newton solver often requires additional load steps to converge to the solution for the prescribed boundary conditions. In contrast, the pre-trained iFOL directly predicts the final solution with acceptable accuracy. From top to bottom, one can observe that by initializing the Newton solver with the response of the DL model, the solution reaches the same level of accuracy as NFEM within only a few iterations. It is worth noting that the number of iterations and thus the speed-up strongly depends on the input sample topology and resolution. This aspect is analyzed in more detail from a statistical perspective later in the paper. 


\begin{figure}[H]
  \centering
\includegraphics[width=0.99\linewidth]{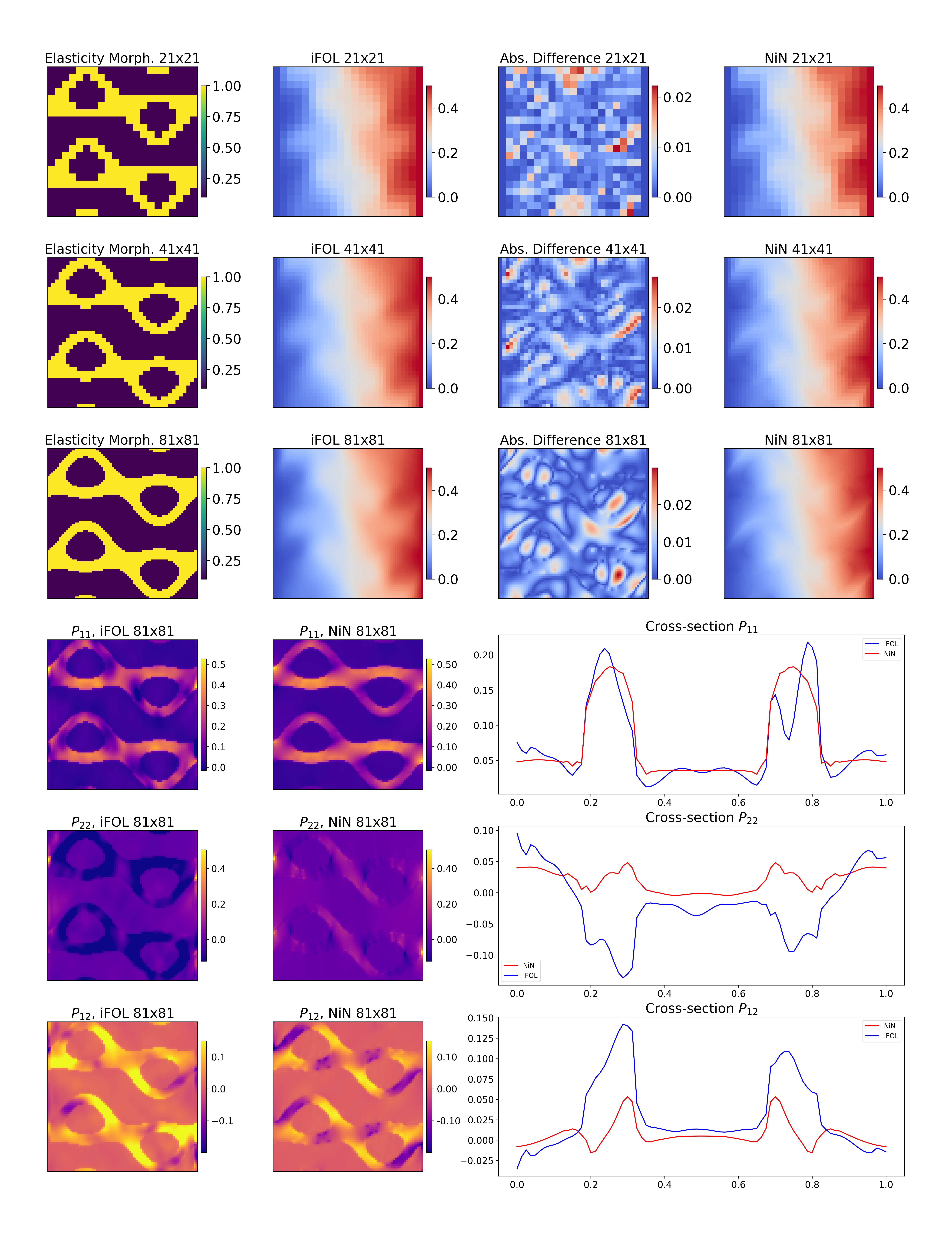}
  \caption{Comparison of iFOL and NiN performance on an unseen metamaterial with complex topology.}
\label{fig:plot_paper_res_81_249}
\end{figure}

Figure~\ref{fig:plot_paper_res_81_249} shows the results for another input sample at three resolutions. The NiN approach, shown in the fourth column, converges reliably using the initialization from the second column. 
We also aim to further discuss the accuracy of the neural operators in predicting spatial derivative fields such as stress components. These quantities are generally more challenging to capture accurately when the network is not directly trained on their supervised values. One major application of the NiN approach is therefore to correct small fluctuations in the primary field, leading to improved approximations of the spatial derivatives.
Note also that, due to the chosen boundary conditions, the deformation components in the $y$-direction are less active and thus take smaller values, which are again harder for the network to capture. In this regard, the NiN approach proves particularly useful for correcting these values.

\begin{figure}[H]
  \centering
  \includegraphics[width=0.99\linewidth]{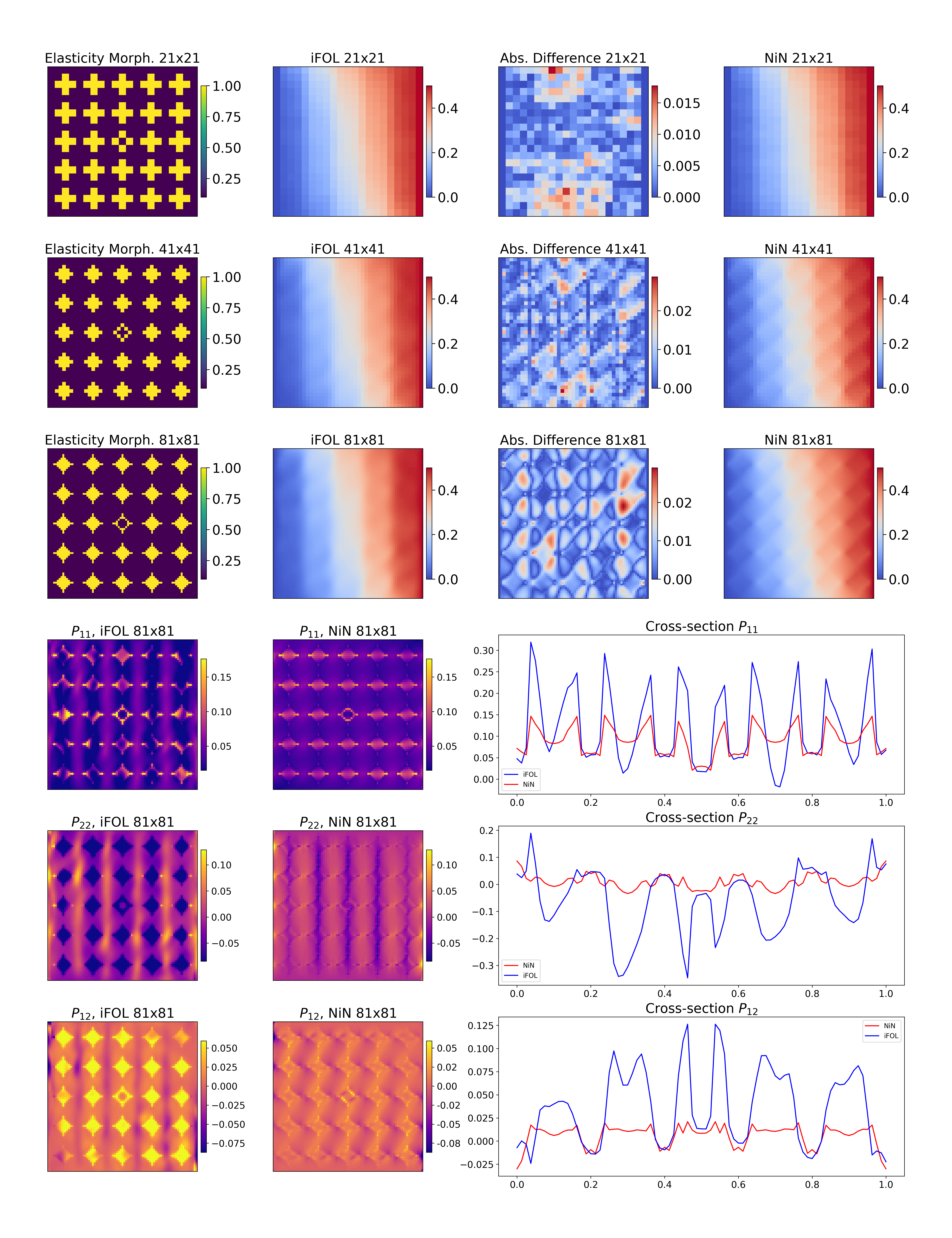}
  \caption{Comparison of iFOL and NiN performance on an unseen microstructure with high-frequency features.  }
\label{fig:plot_paper_res_81_269}
\end{figure}

The same trends hold for another sample provided in Fig.~\ref{fig:plot_paper_res_81_269}. Here, we include hard inclusions with high numbers. The signature of the heterogeneity map is still visible in the displacement field; however, the network smooths these features. This behavior is largely because the network is not exposed to such patterns. However, the predictions are sufficiently accurate for the NiN approach.

\begin{figure}[H]
  \centering
  \includegraphics[width=0.99\linewidth]{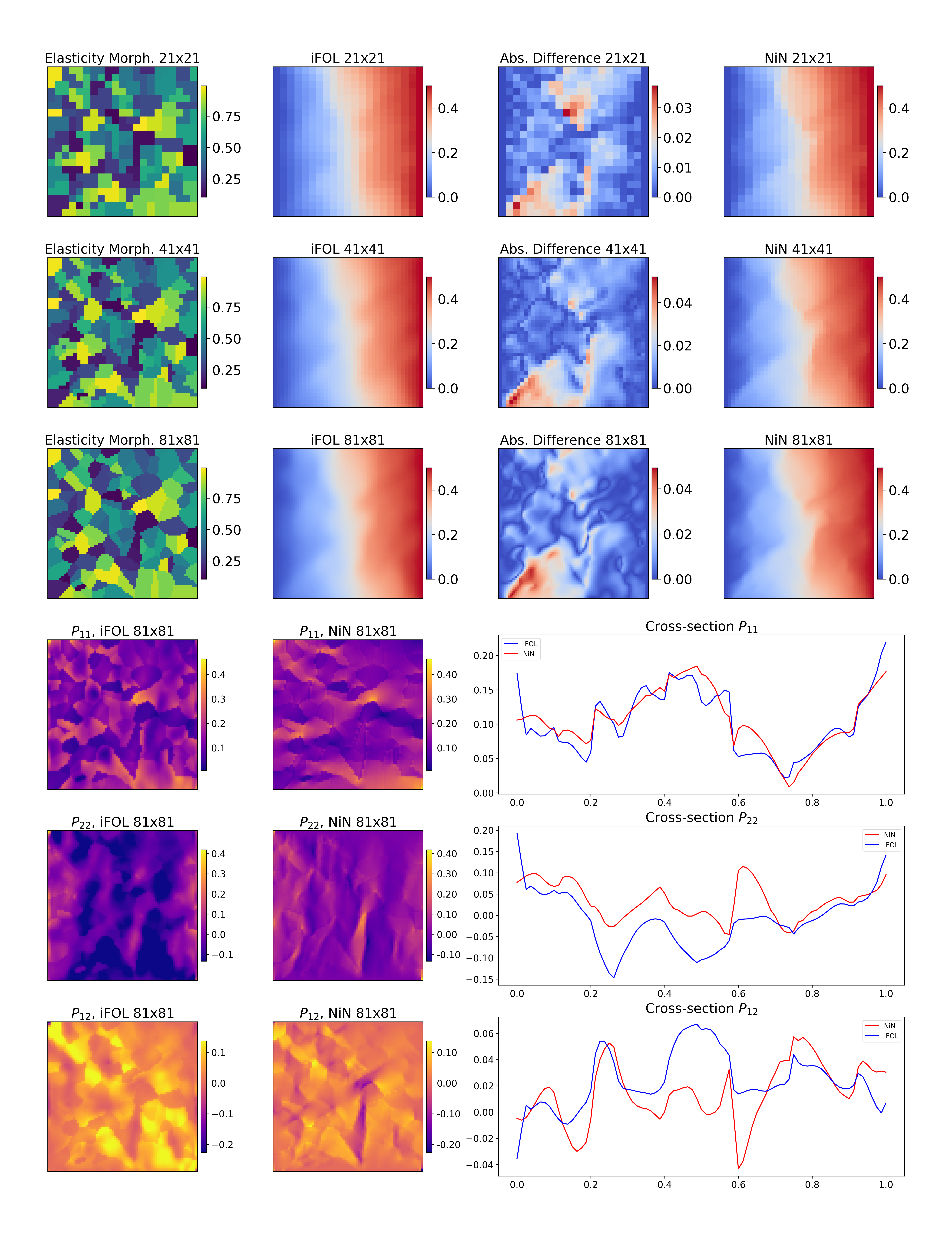}
  \caption{Comparison of iFOL and NiN performance on an unseen polycrystalline microstructure with varying property values.}
\label{fig:plot_paper_res_81_327}
\end{figure}

Very similar conclusions hold for the sample with a polycrystalline topology shown in Fig.~\ref{fig:plot_paper_res_81_327}. In this case, the network faces a more challenging test, as the elastic property values no longer follow the same ratios as in training. Moreover, the overall topology of the property distribution deviates significantly from the training set, leading to potential errors in the iFOL predictions, particularly for the stress components. Once again, the NiN approach can be employed to fine-tune and improve the results.



\begin{figure}[H]
  \centering
\includegraphics[width=0.99\linewidth]{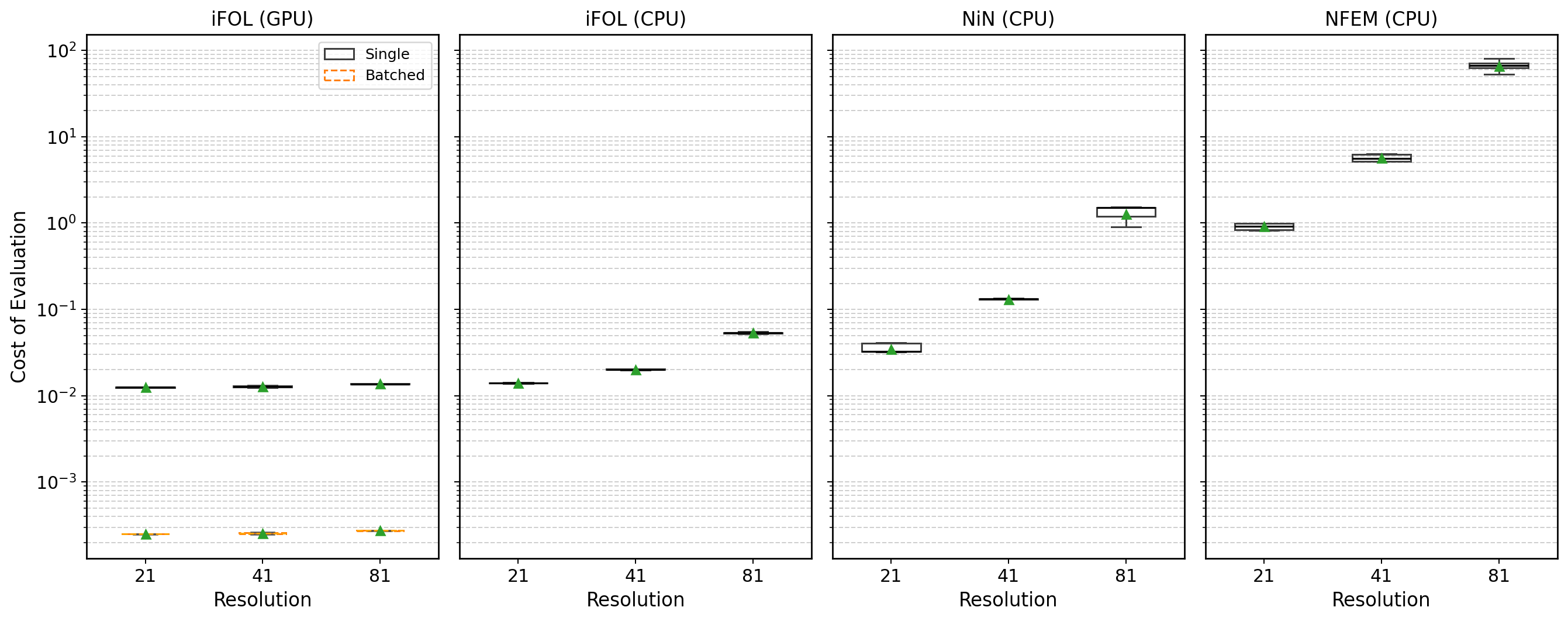}
  \caption{ Comparison of computational cost (in seconds) for iFOL, NiN, and NFEM. NiN achieves the same accuracy as NFEM while being faster, thanks to its proper initialization from iFOL.  }
\label{fig:Cost_eval_box_plot}
\end{figure}

\newpage
In Fig.~\ref{fig:Cost_eval_box_plot}, we compare the computational cost of the three solution strategies. Such a comparison is not straightforward, as several hidden factors influence the final outcome. In addition, aspects such as mesh resolution, degree of nonlinearity, problem dimensionality, and details of the hardware setup, including whether computations are performed on CPUs or GPUs, play a major role. Therefore, reporting a single speed-up factor can be misleading and requires further clarification. In the following, we discuss these factors in more detail.

We start with the previously developed deep learning–based model. As expected, iFOL exhibits the lowest cost at all resolutions since it requires only network inference. No additional cost arises from data generation, as the network is trained directly on the parametric equations (i.e., finite element residuals) without using labeled data. On the other hand, one should consider the training cost, which is intentionally excluded here, as it is moderate (ranging from few hours to a day on a single GPU machine) and typically regarded as a one-time investment (see also Section~\ref{sec:hyperparameters} for more details on the training process). Note that all test samples can be processed in parallel, and predictions can be obtained at any desired resolution. This parallelization further reduces the average inference time per sample, as shown on the far left side of Fig.~\ref{fig:Cost_eval_box_plot}. 

At the other extreme, the NFEM approach exhibits the highest computational cost, which increases sharply with mesh refinement due to the growing number of elements. For a fair comparison, all NFEM simulations were performed on the same CPU system as the other methods, and the reported timings represent statistical averages over all test samples. It is important to note that these timings can vary depending on the implementation details, such as the choice and efficiency of the linear iterative solver. For the current 2D benchmark, we report the best-case scenarios; these values may differ for other examples, as discussed later. Regardless of the specific NFEM implementation, it should be emphasized that the NiN strategy primarily improves the initialization of the nonlinear solver. Therefore, even if the absolute runtimes vary between systems, the relative performance and observed speed-up factors are expected to remain consistent.

The NiN strategy lies between these two: its cost consists of the iFOL inference plus a few Newton iterations in one load step. Although more expensive than iFOL alone, NiN still achieves a one to two orders of magnitude speed-up compared to NFEM.

\newpage
\subsection{Studies on 3D hyper elastic: virtual biaxial testing machine}
\label{sec:3D_elas} 
\begin{figure}[H]
  \centering
\includegraphics[width=0.99\linewidth]{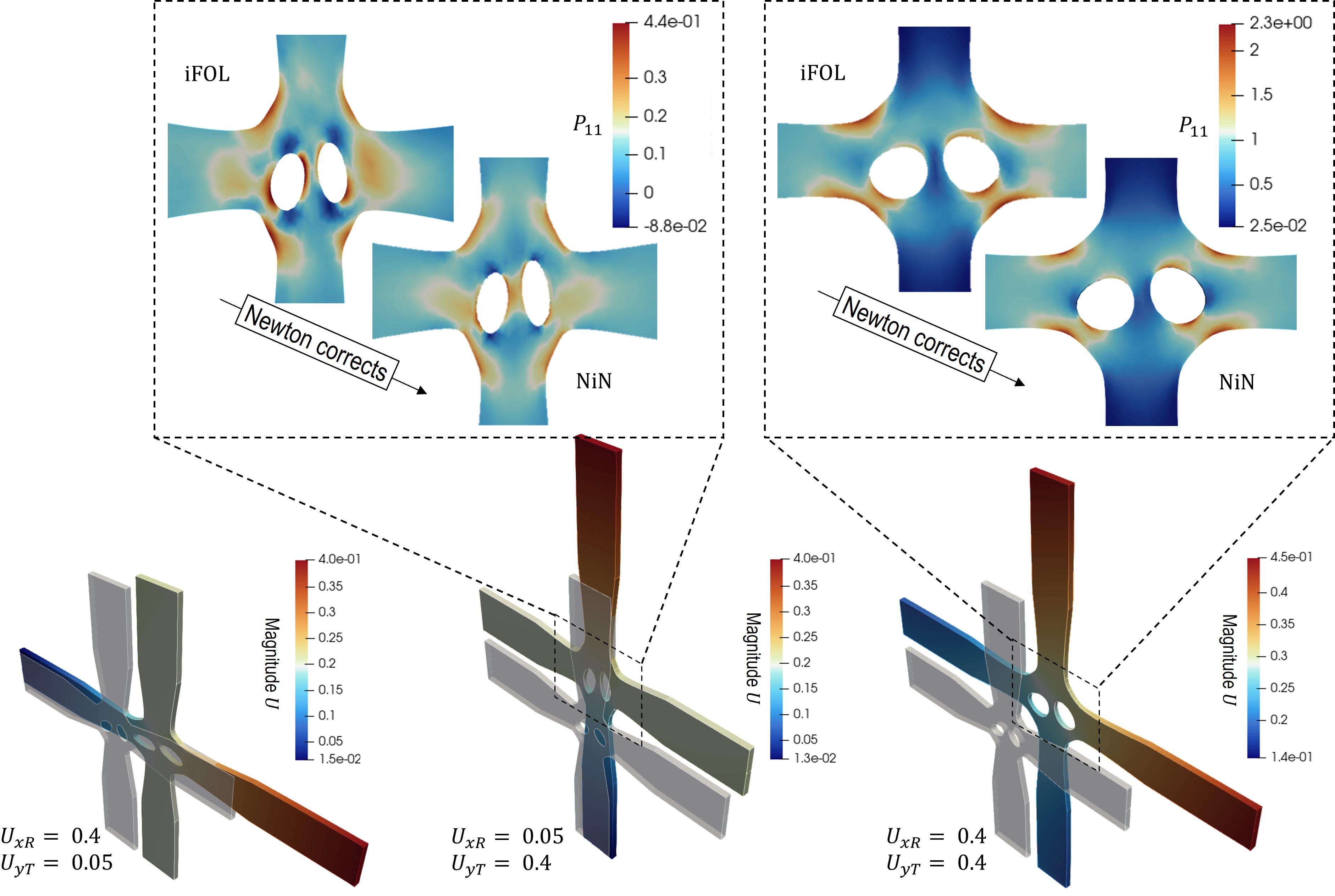}
  \caption{ 3D nonlinear hyperelastic biaxial test: the NiN approach corrects deformation and stress profiles from the neural operator solution on a complex geometry with unseen boundary conditions. }
\label{fig:3D_biaxial}
\end{figure}

One advantage of the iFOL method is its flexibility in handling complex geometries. As a new test, we consider a structural problem: a biaxial loading machine. As shown in the top-left of Fig.~\ref{fig:3D_biaxial}, the setup consists of a 3D cross-shaped structure with two arbitrary holes. The boundary conditions are illustrated on the right and top edges, where different loadings are applied. The goal is to learn the mapping between the applied BCs and the resulting solution field. 
See also Section \ref{sec:formulation_mech3d} and Figs.~\ref{fig:iFOL_3D_mech} and \ref{fig:DBC_dist}
for further details on the formulation and problem setup.

In the second row of Fig.~\ref{fig:3D_biaxial}, three unseen test cases from the NiN are presented, where the NFEM solver is initialized with the results of the pre-trained physics-informed iFOL framework. Due to the anisotropy introduced by the hole placements, we investigate the loading along the two principal axes as well as one case with equal biaxial loading. For comparison, the stress profiles are shown in the top-right plots, where we observe generally good agreement, though not perfect. The NiN successfully refines the solution field, correcting the overprediction of stress components made by the pre-trained neural operator. Compared to NFEM, the NiN approach achieves the results approximately five times faster on average across the test cases shown.

Again, we should emphasize that the deformation patterns originating from the primary field are well captured by the trained neural operator framework (iFOL). On the other hand, the stress profiles, particularly around corners and holes where higher gradients and sensitivities are expected, can be further refined, which is successfully achieved by NiN.

\newpage
\subsection{Studies on 3D hyper elastic: 3D meta material}
\label{sec:3D_metamat} 
Similar to the previous section, we now test the approach on a different problem setting involving a complex-shaped metamaterial subjected to various loading conditions. The training is again performed to learn the deformation and corresponding stress fields for prescribed Dirichlet boundary conditions. The key difference compared to the previous case is that the iFOL model is trained in a supervised manner, using offline NFEM calculations on the given geometry. 

\begin{figure}[H]
  \centering
\includegraphics[width=0.99\linewidth]{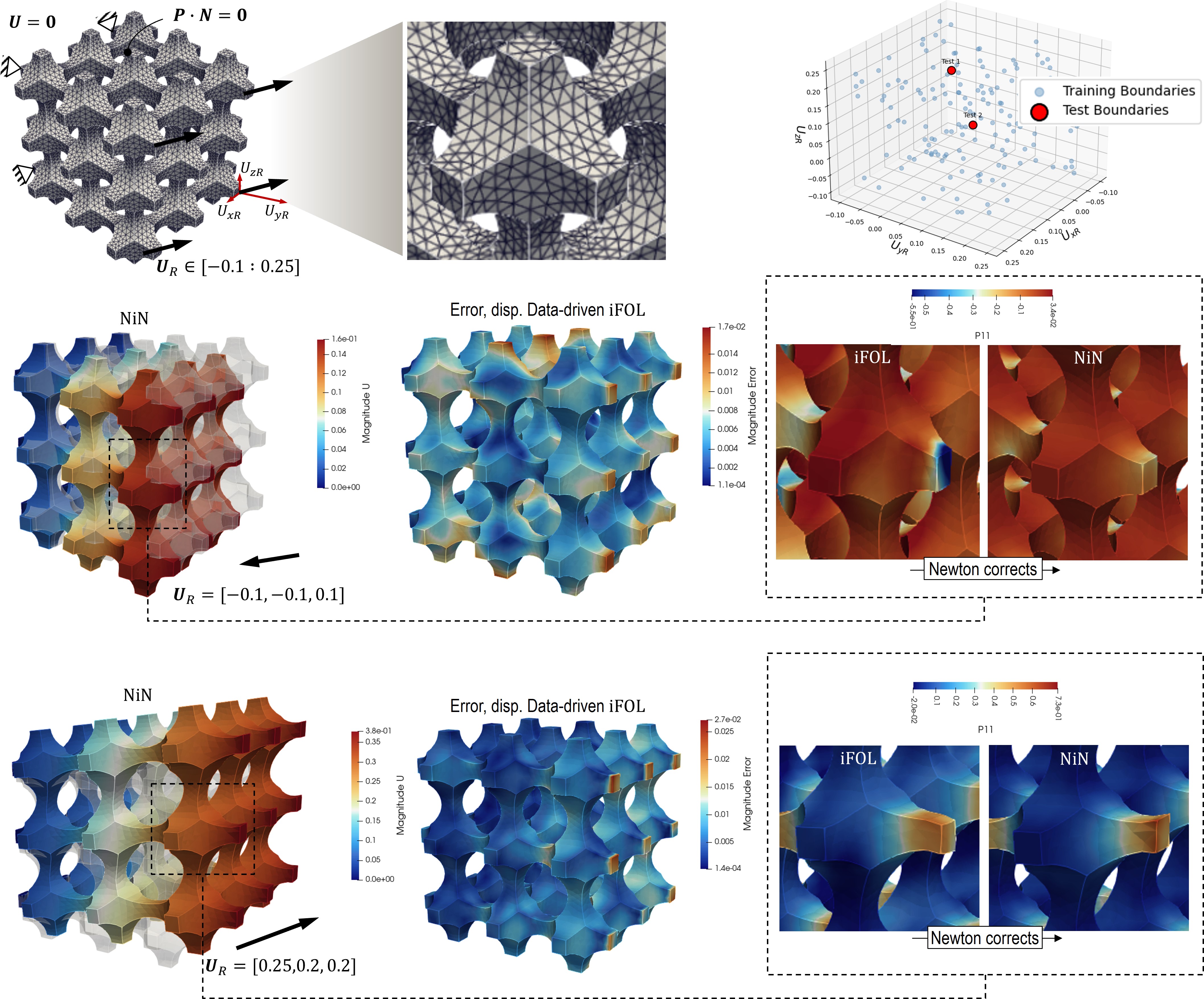}
  \caption{ 3D nonlinear hyperelastic deformation of a meta material with complex shape. }
\label{fig:3D_metamat}
\end{figure}

The first row of Fig.~\ref{fig:3D_metamat} shows the geometry and the distribution of the displacement components for the training and selected test cases. The second and third rows present two additional test cases. We report the pointwise error of the DL model and illustrate how NiN corrects the stress and deformation profiles. The results consistently confirm the effectiveness of the model, reducing the average number of iterations from about 60 to just 5. 

The trained model does not explicitly enforce physical conditions, which is reflected in the higher errors near the boundaries. However, the NiN approach successfully corrects the data-driven DL solution, yielding results that match the NFEM reference and fully respect the underlying physical laws.

\newpage
\subsection{Studies on 3D multiphysics}
\label{sec:3D_thermomech} 

\begin{figure}[H]
  \centering
  \includegraphics[width=0.99\linewidth]{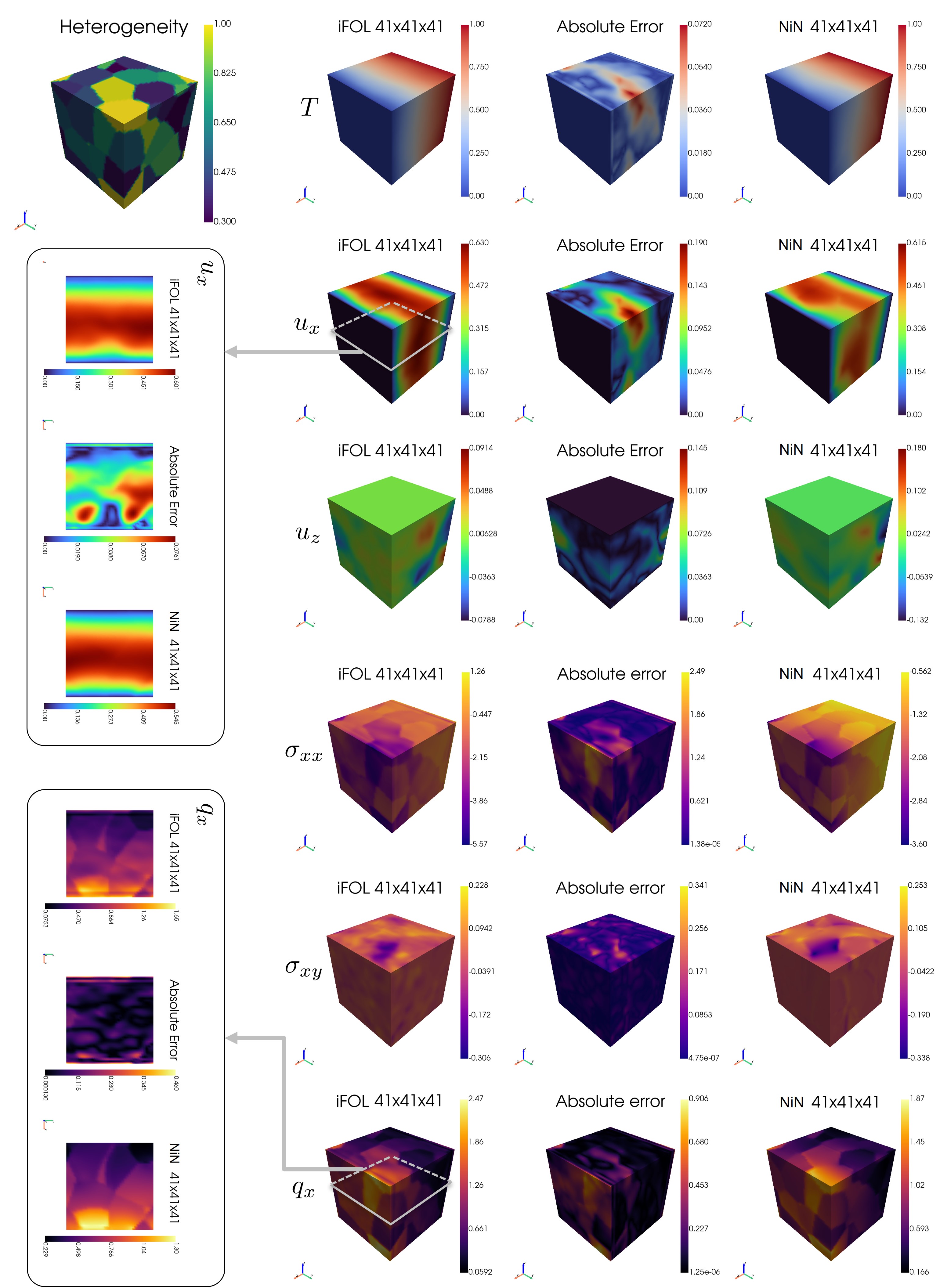}
  \caption{ iFOL performance on an out-of-distribution polycrystalline microstructure and its correction by NiN. }
\label{fig:3D_multi_poly}
\end{figure}
\begin{figure}[H]
  \centering
  \includegraphics[width=0.99\linewidth]{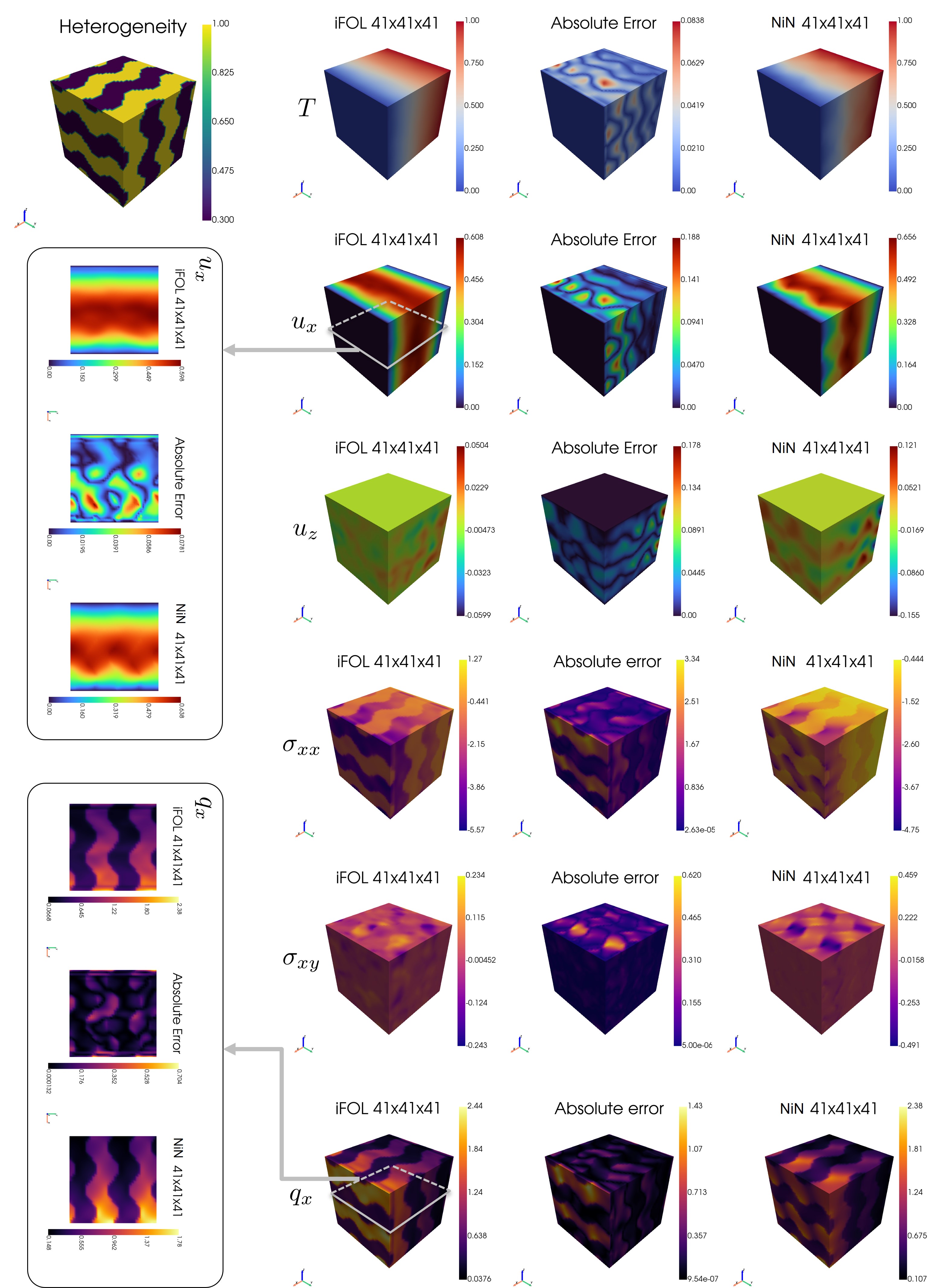}
  \caption{iFOL performance on an out-of-distribution Gyroid-based microstructure and its correction by NiN.}
\label{fig:multi_tmps}
\end{figure}

In the final example, we evaluate the performance of the models for a nonlinear thermomechanical coupled system. See Section \ref{sec:formulation_thermomech} as well as Figs.~\ref{fig:iFOL_3D_thermomech} and \ref{fig:train_test_3D_multi}.
Two extreme test cases are shown in Figs.~\ref{fig:3D_multi_poly} and \ref{fig:multi_tmps}. In each figure, the top-left panel shows the input topology, while the left-hand column of each row presents a selection of the output fields. Due to the 3D nature of the problem, displaying all components of deformation, heat fluxes, and the corresponding stress values would be too extensive and is therefore omitted. Instead, we focus on selected components that exhibit sufficient variation in values and distribution. In addition, two cross sections are presented for the $u_x$ and $q_{x}$ components.

\begin{figure}[H]
  \centering
  \includegraphics[width=0.7\linewidth]{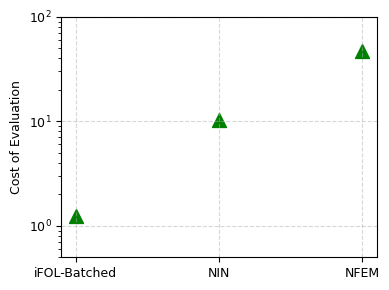}
  \caption{ Comparison of computational cost (in seconds) for iFOL, NiN, NFEM for the 3D nonlinear multiphysics problem. }
\label{fig:Cost_eval_transient}
\end{figure}

The overall trend of the results for the 3D nonlinear multiphysics problem is consistent with the 2D hyperelastic case. Specifically, the iFOL method shows reasonable accuracy for in-distribution scenarios. Still, its performance degrades for out-of-distribution cases, particularly for spatial gradients such as heat flux and stress components. The degradation of results for out-of-distribution samples in the 3D setup is more pronounced than in the 2D case, which is reasonable given the larger degrees of freedom, greater variation in the input samples, and the substantially larger network required for a comparable task. In contrast, the NiN or hybrid approach converges to the solution within only 2 to 3 iterations when initialized, as shown in the second column. This reduction in the number of iterations leads to a lower overall computational cost, as reported in Fig.~\ref{fig:Cost_eval_transient}. 
It is noted that the bi-conjugate gradient stabilized method (BiCGSTAB) is employed as a linear solver in the Newton-Raphson update for the sake of efficiency on large-scale problems. This results in a small speedup compared to the measurements on the 2D hyperelastic example in Fig. \ref{fig:Cost_eval_box_plot}, where the direct linear solver is utilized to solve a linear system of equations.

\newpage
\section{Conclusions and outlook}
\label{sec:conclusions}
We have presented a strategy coined as neural initialized Newton (NiN) for nonlinear finite element problems, where predictions from a pre-trained neural operator (informed by physics or data) are used as the initial guess in the Newton–Raphson solver. We investigated this approach in 2D hyperelastic heterogeneous materials, a 3D structural problem, 3D metamaterials with complex geometries, and 3D nonlinear thermomechanical coupling within a heterogeneous representative volume element. Across all cases, the proposed method consistently reduced the number of Newton iterations and thus the overall computational cost, while retaining the accuracy of the finite element method.

In addition, we exploit the zero-shot super-resolution capability of neural operators. Specifically, the model is trained on relatively low-resolution samples, where training is faster and variability is easier to handle, and then applied at higher resolutions for inference. Although this may introduce prediction errors, the Newton–Raphson solver rapidly corrects them, suggesting that neural operators can act as an effective form of model reduction in nonlinear settings.

\begin{figure}[H]
  \centering
\includegraphics[width=0.8\textwidth]{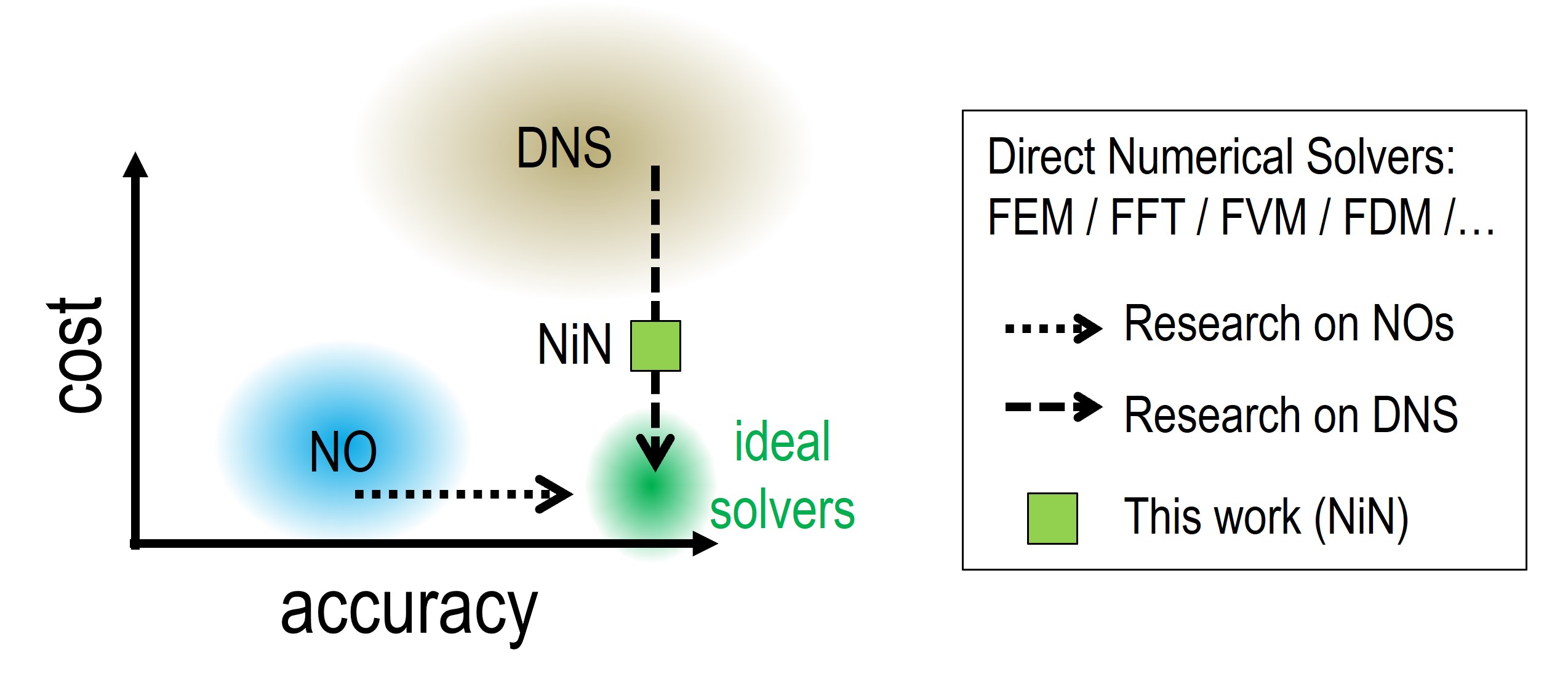}
  \caption{ In the pursuit of an ultimate solver that is both fast and accurate, much of the literature has focused on improving neural operators. Here, we open a new avenue with hybrid approaches that combine the benefits of both realms. In this context, the choice of the direct numerical solver can remain flexible, depending on the application. }
  \label{fig:concl}
\end{figure}

By combining the rapid inference of neural operators at arbitrary resolution with the reliability of classical solvers, NiN provides a balanced solution strategy that mitigates the limitations of each method when used alone. These results highlight the potential of hybrid approaches to accelerate large-scale nonlinear simulations and open promising avenues for future research (see also Fig.~\ref{fig:concl}).


\textbf{Outlook.}
While this study focused on nonlinear and yet path-independent problems in solid mechanics, the framework can, in principle, be extended to path-dependent PDEs, spatiotemporal problems, and other areas of computational mechanics \cite{yamazaki2024, NAJAFIKOOPAS2025110675, MALLEVAL2025104431}.
Although beyond the present scope, the same idea may also be extended to other nonlinear solvers, such as FFT-based methods (see \cite{HARANDI2025106219} for related discussions and references therein). It should be emphasized that the concepts presented in this work are not limited to iFOL and could easily be coupled with alternative neural operator formulations.
\\ \\ \\ \\
\noindent
\textbf{Data Availability}:
All data and code used for this study are publicly available in the open repository at \href{https://github.com/Harandi-Ali/SPiFOL}{iFOL}. 
\\ \\
\textbf{Acknowledgements}:
K.T. and Sh.R. appreciate the Deutsche Forschungsgemeinschaft (DFG) for the funding support provided to develop the present work in the project 561202254.
Sh.R. and M.A. would like to thank DFG for the funding support in the project Cluster of Excellence Internet of Production (project: 390621612). The authors gratefully acknowledge the computational resources provided by the WestAI GPU cluster under project RWTH1870. Y.Y. acknowledges the funding support by JST BOOST, Japan Grant Number JPMJBS2409.
\\ \\ 
\textbf{Author Statement}:
K.T.: Methodology, Software, Writing - Review \& Editing. 
J.P.V.: Software, Review \& Editing. 
Y.Y.: Software, Methodology,  Writing - Review \& Editing. 
M.A.: Funding, Supervision, Review \& Editing.
R.N.A.: Methodology, Software, Review \& Editing. 
Sh.R.: Conceptualization, Methodology, Writing - Review, Funding, Supervision \& Editing.
\\ 
\appendix

\newpage
\section{Sample generation}
\label{sec:data}
\color{black}
One of the main challenges in operator learning is ensuring that the trained model generalizes well across a broad parametric space, including variations in microstructure geometry, boundary conditions, and other problem-specific quantities. A sufficiently diverse dataset is therefore essential. This appendix provides a detailed analysis of the generated microstructure samples used during training and testing; see also Sections~\ref{sec:2D_elas} and~\ref{sec:3D_thermomech}.

In particular, we focus on a parametrization we refer to as Fourier-based, which has shown promising performance in our previous studies \cite{Rezaei2024fol_mech, yamazaki2024}. Although we do not provide a formal proof, empirical evidence suggests that this representation improves the ability of the network to generalize to unseen microstructures. The underlying motivation stems from the observation that any arbitrary topology can be expressed as a superposition of Fourier modes with different frequencies. Training the physics-informed operator learning model with such fundamental components appears to enhance extrapolation to unseen cases.

In two dimensions, the Fourier-based parametrization of the heterogeneous microstructure is represented by a combination of sine and cosine functions of the form:
\begin{equation}
\label{eq:foruier_eq}
\begin{aligned}
    \phi_f(x,y) &= \sum_i^{n_{sum}} [c_i + A_i \sin{(f_{x,i}~x)} \cos{(f_{y,i}~y)} + B_i \cos{(f_{x,i}~x)} \sin{(f_{y,i}~y)} \\ & + C_i\sin{(f_{x,i}~x)} \sin{(f_{y,i}~y)} + D_i \cos{(f_{x,i}~x)} \cos({f_{y,i}~y)}].
\end{aligned}
\end{equation}

Here, $c_i$ denotes a real-valued constant, while $\{A_i, B_i, C_i, D_i\}$ are the amplitudes associated with the corresponding frequency mode. The quantities $\{f_{x,i}, f_{y,i}\}$ represent the frequencies in the $x$- and $y$-directions, respectively.  
To enhance realism and generate more intricate microstructural patterns, we further introduce a phase function $\phi_f$, obtained through a sigmoidal projection:

\begin{equation}
\label{eq:sigmoid}
\begin{aligned}
    \phi(x,y) = (\phi_{max}-\phi_{min})\cdot\text{Sigmoid}\left(\beta(\phi_f-0.5)\right) + \phi_{min}. 
\end{aligned}
\end{equation}
The above projection ensures that the phase values remain bounded between $\phi_{\min}$ and $\phi_{\max}$. The parameter $\beta$ controls the sharpness of the transition between the two phases.  
For simplicity, in the subsequent analysis, we retain only the constant term together with the final term involving the product of two cosine functions, i.e., $A_i = B_i = C_i = 0$.

For the 2D studies in section \ref{sec:2D_elas}, we select three frequencies for each direction. Considering the constant term, this choice gives us $M = 3 \times 3 + 1 = 10$ different terms, which can be added to construct $\phi(x,y)$ according to Eq.~\ref{eq:sigmoid} and Eq.~\ref{eq:foruier_eq}.

For all heterogeneous materials, properties in the presented studies are defined through the phase function $\phi(\boldsymbol{X})$. For example, in Sections~\ref{sec:2D_elas} and \ref{sec:3D_thermomech}, the elastic properties are prescribed as the spatially varying Young’s modulus $E(\boldsymbol{X}) = \phi(\boldsymbol{X})$. Consequently, the minimum and maximum stiffness values are normalized according to the range of $\phi(\boldsymbol{X})$. For test cases such as polycrystalline structures or other microstructures with varying phase contrasts, the distribution of $E(\boldsymbol{X})$ is selected arbitrarily depending on the application. It is worth noting, however, that during training we restrict ourselves to samples with a fixed phase contrast ratio, i.e., $E_{\max}/E_{\min}$.
Extension of the above formulation to 3D is straightforward (see also \cite{HARANDI2025106219}). 



The input samples used to train the model in Section \ref{sec:2D_elas} are generated by combining different random distributions for these ten inputs. See also Table \ref{tab:Data_gen_2D} for a summary of Fourier-based parameterization. For the samples used in Section \ref{sec:3D_thermomech}, see Table \ref{tab:Data_gen_3D}

\begin{table}[H]
\centering
\caption{Data generation parameters for 2D Fourier-based approach}  
\label{tab:Data_gen_2D}
\begin{footnotesize}
\begin{tabular}{ l l l l l }
\hline
     Total num. of samples & parameters &  values & $\beta$ & $E_{\max}/E_{\min}$ \\
\hline
8000 & $\left( \bm{f}_x,~\bm{f}_y\right)$ &  $\left(\left[0,\,1,\,2,\,3\right],~\left[0,\,1,\,2,\,3\right] \right)$ 
&  $20$ & $1.0/0.1$ \\
\hline
\end{tabular}
\end{footnotesize}
\end{table} 

\begin{table}[H]
\centering
\caption{Data generation parameters for 3D Fourier-based approach}  
\label{tab:Data_gen_3D}
\begin{footnotesize}
\begin{tabular}{ l l l l l }
\hline
     Num. of samples & parameters &  values & $\beta$  & $E_{\max}/E_{\min}$ \\
\hline
1000 & $\left( \bm{f}_x,~\bm{f}_y,~\bm{f}_z\right)$ &  $\left(\left[1,\,2,\,3\right],~\left[1,\,2,\,3\right],~\left[1,\,2,\,3\right] \right)$ 
&  $10$  & $1.0/0.3$ \\
1000 & $\left( \bm{f}_x,~\bm{f}_y,~\bm{f}_z\right)$ &  $\left(\left[1,\,2\right],~\left[1,\,2\right],~\left[1,\,2\right] \right)$ &  $10$ &  $1.0/0.3$  \\
1000 & $\left( \bm{f}_x,~\bm{f}_y,~\bm{f}_z\right)$ &  $\left(\left[1,\,2,\,4,\,8\right],~\left[1,\,2,\,4,\,8\right],~\left[1,\,2,\,4,\,8\right] \right)$ 
&  $10$  & $1.0/0.3$\\
1000 & $\left( \bm{f}_x,~\bm{f}_y,~\bm{f}_z\right)$ &  $\left(\left[2,\,4,\,6\right],~\left[2,\,4,\,6\right],~\left[2,\,4,\,6\right] \right)$ 
&  $10$  & $1.0/0.3$\\
1000 & $\left( \bm{f}_x,~\bm{f}_y,~\bm{f}_z\right)$ &  $\left(\left[1,\,3,\,5,\,7\right],~\left[1,\,3,\,5,\,7\right],~\left[1,\,3,\,5,\,7\right] \right)$ 
&  $10$  & $1.0/0.3$\\
\hline
\end{tabular}
\end{footnotesize}
\end{table} 

\begin{figure}[H]
  \centering
\includegraphics[width=0.8\textwidth]{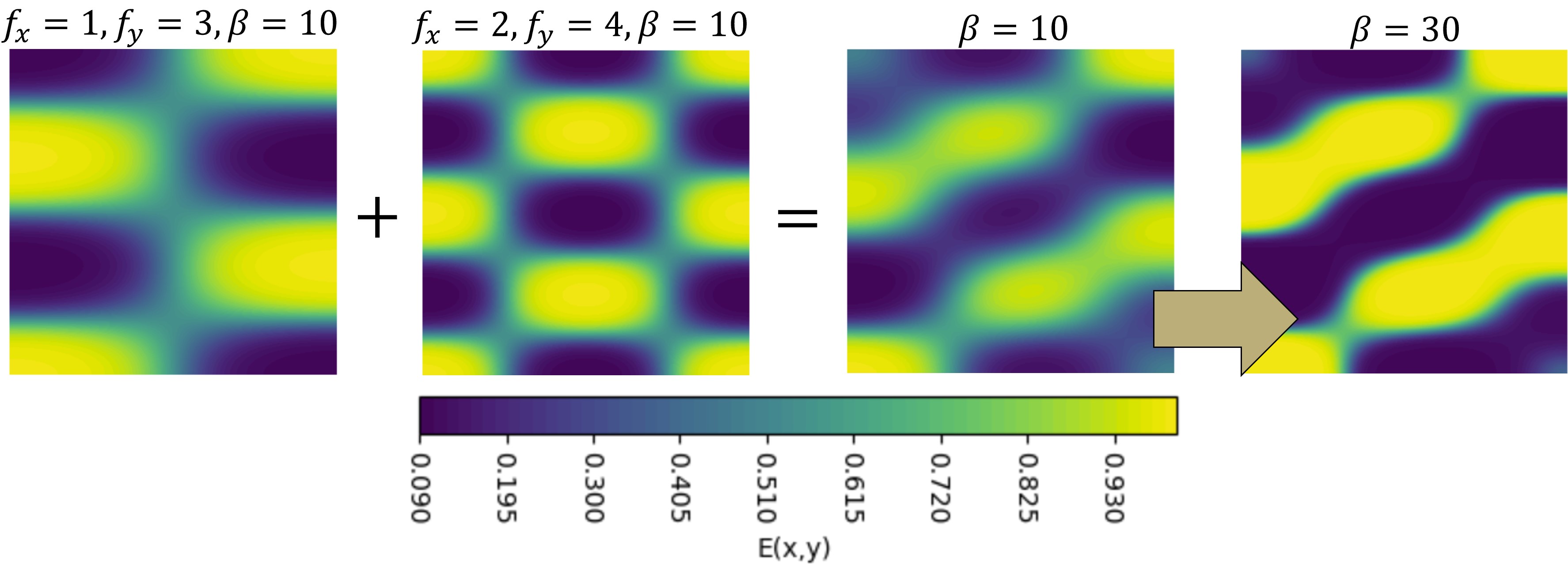}
  \caption{ Illustration of how Fourier-based parameterization can help construct different topologies through simple combinations of multiple frequencies.}
  \label{fig:fp}
\end{figure}

For illustration, Fig.~\ref{fig:fp} demonstrates how different frequency components can be combined to generate complex topologies. Periodic and symmetric patterns, for instance, can be generated using even frequency numbers. On the right-hand side, the influence of the parameter $\beta$ on the sharpness of the phase transition zone is also illustrated. More details can be found in \cite{Rezaei2024finite, Rezaei2024fol_mech}.


\section{A short review on FEM formulation}
\label{sec:review_FEM}
We briefly summarize the FEM discretization for a 2D quadrilateral element to clarify the loss terms in the iFOL formulation. The procedure is standard, and extensions to higher-order or more complex elements follow directly \cite{bathe1996finite}. In this work, we employ the linear shape functions $\boldsymbol{N}$ and the deformation matrix $\boldsymbol{B}$ to discretize the mechanical weak form.

\begin{align}
\label{eq:N}
\bm{N} =
\begin{bmatrix}
N_1 & 0 & \dots & N_4 & 0\\
0 & N_1 & \dots & 0 & N_4
\end{bmatrix},~
\bm{B} =
\begin{bmatrix}
N_{1,x} & 0       & \dots & N_{4,x} & 0        \\
0       & N_{1,y} & \dots & 0       & N_{4,y}  \\
N_{1,y} & N_{1,x} & \dots & N_{4,y} & N_{4,x} 
\end{bmatrix}.
\end{align}
The notation $N_{i,x}$ and $N_{i,y}$ represent the derivatives of the shape function $N_i$ with respect to the coordinates $x$ and $y$, respectively. To compute these derivatives, we utilize the Jacobian matrix
\begin{align}
 \boldsymbol J = \partial \boldsymbol X / \partial \boldsymbol \xi = \begin{bmatrix} \frac{\partial x}{\partial \xi} & \frac{\partial y}{\partial \xi} \\ \frac{\partial x}{\partial \eta} & \frac{\partial y}{\partial \eta} \end{bmatrix}. 
\end{align}

Here, $\boldsymbol X = [x, y]$ and $\boldsymbol \xi = [\xi, \eta]$ represent the physical and parent coordinate systems, respectively. It is worth mentioning that this determinant remains constant for parallelogram-shaped elements, eliminating the need to evaluate this term at each integration point.
Finally, for the B matrix in linear case, we have 
\begin{align}
 \bm{B} = \bm{J}^{-1} \begin{bmatrix} \frac{\partial N_1}{\partial \xi} & 0 & \frac{\partial N_2}{\partial \xi} & 0 & \frac{\partial N_3}{\partial \xi} & 0 & \frac{\partial N_4}{\partial \xi} & 0 \\ 0 & \frac{\partial N_1}{\partial \eta} & 0 & \frac{\partial N_2}{\partial \eta} & 0 & \frac{\partial N_3}{\partial \eta} & 0 & \frac{\partial N_4}{\partial \eta} \\ \frac{\partial N_1}{\partial \eta} & \frac{\partial N_1}{\partial \xi} & \frac{\partial N_2}{\partial \eta} & \frac{\partial N_2}{\partial \xi} & \frac{\partial N_3}{\partial \eta} & \frac{\partial N_3}{\partial \xi} & \frac{\partial N_4}{\partial \eta} & \frac{\partial N_4}{\partial \xi} \end{bmatrix}. 
\end{align}

For each element the deformation field and stress tensor $\hat{\bm{\sigma}}$ are approximated as
\begin{align}
\boldsymbol{u}_e =\boldsymbol{N} \boldsymbol U_e ,~~~~
\hat{\bm{\sigma}} &= \bm{C}\boldsymbol{B} \bm{U}_e, 
\end{align}
Here, $\boldsymbol U^T_e=\left[
\begin{matrix}
U_{1,x}  &\cdots & U_{4,y} \\
\end{matrix}
\right]$ is the nodal values of the deformation field of element $e$. 
The same procedure applies to the thermal problem, and the 3D extension is also trivial, based on the standard FE procedure.


In the nonlinear hyperelastic examples, the following general form of the $\boldsymbol{B}^{nl}$ matrix was used.

\begin{equation}
\bm{B}^{nl}_{I} =
\begin{bmatrix}
F_{11}\,\frac{\partial N_I}{\partial X} & F_{21}\,\frac{\partial N_I}{\partial X} 
\\[6pt]
F_{12}\,\frac{\partial N_I}{\partial Y} & F_{22}\,\frac{\partial N_I}{\partial Y} 
\\[6pt]
F_{11}\,\frac{\partial N_I}{\partial Y} + F_{12}\,\frac{\partial N_I}{\partial X} &
F_{21}\,\frac{\partial N_I}{\partial Y} + F_{22}\,\frac{\partial N_I}{\partial X} 
\end{bmatrix}.
\end{equation}

The matrix $\boldsymbol{B}^{\mathrm{nl}}$ is used to construct the finite element approximation of $\delta \hat{\boldsymbol{ E}}$, through which the residual in each element can be defined as follows:

\begin{equation}
\delta \hat{\bm{E}} \approx 
\begin{bmatrix}
\bm{B}_1^{e} & \bm{B}_2^{e} & \cdots & \bm{B}_{n_e}^{e}
\end{bmatrix}_{3 \times 2n_e}
\begin{Bmatrix}
\delta \bm{u}_1^{e} \\[3pt]
\delta \bm{u}_2^{e} \\[3pt]
\vdots \\[3pt]
\delta \bm{u}_{n_e}^{e}
\end{Bmatrix}_{2n_e \times 1}
= \bm{B}^{nl} \, \delta \bm{u}^{e},
\end{equation}


The second Piola–Kirchhoff stress tensor is obtained as the derivative of the corresponding hyperelastic strain energy function $W$ with respect to the right Cauchy–Green deformation tensor $\boldsymbol{C}$, as follows.

\begin{equation}
    \bm{S} = 2\frac{\partial{W}}{\partial{\bm{C}}}
\end{equation}

\color{black}


\newpage

\section{Hyperparameters and hardware details}
\label{sec:hyperparameters}

A comprehensive summary of the network's hyperparameters and configurations for all the models introduced above, as well as the various analyses conducted, is provided in Table~\ref{tab:hyperparam_stan}. NVIDIA Quadro RTX 6000s with 24 GB of RAM utilized for the training of the 2D hyper elastic model and NVIDIA H100 GPU is used for training of the 3D mechanical models. The studies on the multiphysics problems were performed on NVIDIA GeForce RTX 4090 with 24 GB of RAM. Obviously, the training time of each model varies significantly and depends on many factors, such as the desired accuracy level, the number of training fields, the number of epochs, and other hyperparameters. Depending on the model complexity and the computational power of the GPUs used, the training time ranged from about 2 hours (for data-driven 3D cases) to up to 3 days (for 3D physics-informed multiphysics nonlinear problems).

We studied several parameters, such as latent size, learning rate, network structure, and the frequency of sinusoidal functions. The best settings for iFOL are listed in Tables \ref{tab:hyperparam_stan} for each problem. We found that a frequency parameter of \(\omega_0\) other than $30$ worked best in different problems, while other values did not give clear improvement. Increasing network depth and latent size was generally helpful with sinusoidal activation, although this also increased the training cost. For simpler problems, using fewer epochs and latent iterations can speed up training and inference. As shown in the results, using more samples improves accuracy, but this also increases training cost depending on batch size. In practice, both the network architecture and latent size should be adapted to balance cost and performance.

\begin{table}[H]
\centering
 \caption{Network hyperparameters for stationary problems 
}
 \begin{tabular}{l l l l l}
    \hline
         Training parameter &  Sec.~\ref{sec:2D_elas}
 & Sec.~\ref{sec:3D_elas} & Sec.~\ref{sec:3D_metamat} & Sec.~\ref{sec:3D_thermomech}  \\
    \hline
     Number of samples & 8000 & 800 & 200 & 4000\\    
     Grid in training & $41 \times 41$ & $2454$ & $16593$ & $21 \times 21 \times 21$\\
     Grid in evaluation & $21^2,~41^2,~81^2$ & $2454$ & $16593$ & $21^3,~41^3$\\
     Synthesizers  & [64]*4 & [64]*1 & [64]*4 & [256]*6\\
     \(\omega_0\) & 30 & 30 & 30 & 30\\
     Modulators & Linear (FiLM) & Linear (FiLM) & Linear (FiLM) & Linear (FiLM)\\
     Latent size & 512 & 512 & 64 & 512\\
     Number of latent iterations & 3 & 3 & 3 & 3\\     
     Latent/encoding learning rate & $10^{-2}$ & $10^{-2}$ & $10^{-2}$ & $10^{-2}$\\     
     Training learning rate & $10^{-5}$ & $10^{-5}$ & $10^{-5}$  & $10^{-4}$ to $ 10^{-7}$ \\
     Batch size & 320 & 100 & 5 & 25 \\
     Gradient normalization & Yes & Yes & Yes & Yes\\
     Number of epochs & 10000 & 20000 & 50000 & 10000\\
     Optimizer& Adam & Adam & Adam & Adam\\     
    \hline
    Total trainable parameters & 143938 & 144003 & 29315 & 1117444\\
    \hline\\
    \end{tabular}
    \label{tab:hyperparam_stan}
\end{table}


\newpage
\bibliography{Ref}

\end{document}